\documentclass{article}
\usepackage{arxiv}
\usepackage[utf8]{inputenc}
\usepackage[T1]{fontenc}
\usepackage{hyperref} 
\usepackage{url}
\usepackage{booktabs}
\usepackage{amsfonts}
\usepackage{nicefrac}
\usepackage{microtype}
\def\acks{\subsection*{Acknowledgements}}
\usepackage{natbib}
\usepackage{hyperref}
\usepackage{amsmath,amssymb}
\usepackage{adjustbox}

\usepackage{multirow}
\usepackage{subcaption}
\usepackage{pdfpages}
\usepackage{color}

\DeclareMathOperator{\E}{\mathbb{E}}

\begin{document}

%%%%%% ArXiv %%%%%%%
\title{Pre-training with Non-expert Human Demonstration for Deep Reinforcement Learning}

\author{
  Gabriel V.~de la Cruz Jr., Yunshu Du and Matthew E.~Taylor \\
  School of Electrical Engineering and Computer Science\\
  Washington State University\\
  Pullman, WA 99164-2752 \\
  \texttt{\{gabriel.delacruz,yunshu.du,matthew.e.taylor\}@wsu.edu} \\
}
\maketitle
%%%%%% ArXiv %%%%%%%

% %%%%%%% KER %%%%%%%%
% \KER{1}{24}{00}{0}{2018}{S000000000000000}

% \runningheads{G. V. de la Cruz Jr., Y. Du, and M. E. Taylor}{Pre-training with Non-expert Human Demonstration}

% %\doublespacing

% \title{Pre-training with Non-expert Human Demonstration for Deep Reinforcement Learning}

% \author{GABRIEL V.~DE LA CRUZ JR., YUNSHU DU, and MATTHEW E.~TAYLOR}

% \address{Washington State University, 
% Pullman, Washington 99164-2752, USA \\
% \email{gabriel.delacruz@wsu.edu, yunshu.du@wsu.edu, matthew.e.taylor@wsu.edu}}
% %%%%%%% KER %%%%%%%%

\begin{abstract}
Deep reinforcement learning (deep RL) has achieved superior performance in complex sequential tasks by using deep neural networks as function approximators to learn directly from raw input images. However, learning directly from raw images is data inefficient. The agent must learn feature representation of complex states in addition to learning a policy. As a result, deep RL typically suffers from slow learning speeds and often requires a prohibitively large amount of training time and data to reach reasonable performance, making it inapplicable to real-world settings where data is expensive. In this work, we improve data efficiency in deep RL by addressing one of the two learning goals, feature learning. We leverage supervised learning to pre-train on a small set of non-expert human demonstrations and empirically evaluate our approach using the asynchronous advantage actor-critic algorithms (A3C) in the Atari domain. Our results show significant improvements in learning speed, even when the provided demonstration is noisy and of low quality.
\end{abstract}

\section{Introduction}
The widespread successes of deep Reinforcement Learning (deep RL) have brought a resurgence in using deep neural networks for RL tasks. Using a deep neural network as its function approximator, deep RL can learn state representations directly from raw input pixels and has achieved state-of-the-art results in various domains \citep{mnih2015human, silver2016mastering, kempka2016vizdoom, lillicrap2015continuous, duan2016benchmarking, silver2018general}. However, despite this impressive ability, deep RL remains data inefficient and often requires a long training time to achieve reasonable performance. This drawback has made deep RL impractical in real-world applications where data is expensive to collect, such as in robotics, self-driving cars, finance, or medical applications \citep{bojarski2017explaining, deng2017deep, miotto2017deep}.

Similar to classic RL algorithms, deep RL suffers from poor initial performance since it learns \emph{tabula rasa} \citep{sutton2018reinforcement}. Inherently, deep RL takes even longer to learn because, unlike in classic RL where hand-engineered features were used, deep RL has to learn features directly from raw observations, in addition to policy learning. Therefore, one can speed up deep RL by addressing its two learning components: feature learning and policy learning. In this work, we tackle the feature learning problem and show that faster learning can be achieved by addressing one of the two problems. 

There have been many techniques proposed to speed up feature learning in deep RL, from transfer learning \citep{taylor2009transfer, pan2010survey}, to reward shaping \citep{ng1999policy, brys2015policy}, to using auxiliary tasks \citep{zhang2016augmenting, jaderberg2017unreal, mirowski2017learning, papoudakis18, du2018adapting}. Learning from demonstrations \citep{argall2009survey} is yet another way to help speed up learning in deep RL and has recently gained traction due to its ability to bootstrap the agent at the beginning of training \citep{kurin2017atari,vinyals2017starcraft,hester2018learning, pohlen2018observe}. In this work, we make use of human demonstration data by first using supervised learning to pre-train a neural network to learn the underlying state features and then transfer the learned features to an RL agent \citep{erhan2009difficulty, Erhan:2010:WUP:1756006.1756025,yosinski2014transferable}. We evaluate our approach using a recently-developed deep RL algorithm, the Asynchronous Advantage Actor-Critic (A3C) \citep{mnih2016asynchronous} algorithm in six Atari games \citep{bellemare13arcade}. Unlike previous work where a large amount of expert human data is required to achieve good initial performance boost, our approach shows significant learning speed improvements on all experiments with only a relatively small amount of noisy, non-expert demonstration data. The simplicity of our approach has made it generally adaptable to other deep RL algorithms and potentially to other domains since the collection of demonstration data becomes easy. In addition, we apply Gradient-weighted Class Activation Mapping (Grad-CAM) \citep{selvaraju2017grad} on learned feature maps for both the human data and the agent data, providing a detailed analysis on why pre-training helps to speed up learning. Our work makes the following contributions: 
\begin{enumerate}
    \item We show that pre-training on a small amount of non-expert human demonstration data is sufficient to achieve significant performance improvements. %\yunshu{I think our method did not show "jumpstart" so changing "boost at the start of training" to just "improvements"}
    \item We are the first 
    %(to the best of our knowledge) 
    to apply the transformed Bellman (TB) operator \citep{pohlen2018observe} in the A3C algorithm \citep{mnih2016asynchronous} and further improve A3C's performance on both baseline and pre-training methods. 
    %\item we show pre-training achieves significant improvement when the suboptimality of reward clipping is addressed; \MET{Expand on this point? Not sure what it means} %addressing the issue of reward clipping that our pre-training approach shows significant improvement;
    \item We propose a modified version of the Grad-CAM method \citep{selvaraju2017grad}, which we are the first %(to the best of our knowledge) 
    to provide empirical analysis on what features are learned from pre-training, indicating why pre-training on human demonstration data helps.
    \item We release our code and all collected human demonstration data at \url{https://github.com/gabrieledcjr/DeepRL}.
\end{enumerate}

This article is organized as the following. In the next section, we review some of the related work in using pre-training to improve data efficiency. Section \ref{sec:background} provides background on deep RL algorithms and the transformed Bellman operator. In Section \ref{sec:pretrain}, we propose our pre-training methods for deep RL. Followed by Section \ref{sec:experiment} where we describe the experimental designs. Results and analysis are presented in Section \ref{sec:results}. We conclude this article in Section \ref{sec:conclusion} with discussions and future works.

\section{Related Work}
Our work is closely related to transfer learning \citep{taylor2009transfer, pan2010survey} where learned knowledge from source task(s) is transferred to target task(s) such that the target task does not need to learn features and/or policies from scratch, thus obtaining faster learning. In supervised learning, transferring parameters from a model pre-trained on ImageNet \citep{russakovsky2015imagenet} has shown to be an effective way of speeding up image classification in a new dataset, especially when the source dataset is similar to the target dataset \citep{yosinski2014transferable}. In deep RL, the performance of a target agent can be improved by making use of the knowledge learned in one or more similar source agents \citep{du2016initial, glatt2016, parisotto2015actor, rusu2015policy, teh2017distral}.
All works mentioned above perform pre-train and transfer under the same problem settings. That is, pre-train in supervised learning and transfer to supervised learning, or pre-train in RL and transfer to RL. In this work, we consider a different mechanism that transfers between different problem settings. We pre-train a supervised classification model on the human demonstration data, then transfer the learned features to an RL agent.

Existing work has shown that pre-training can effectively speed up learning in RL. For example, \cite{abtahi2011deep} considered learning latent features with unsupervised learning through Deep Belief Networks as a pre-training method; \cite{anderson2015faster} pre-train hidden units of a Q-network by learning to predict state dynamics. These works show learning speed up in relatively easy RL domains such as Mountain Car, Puddle World, and Cart-Pole. Our approach is in spirit to these earlier works and differs in that we consider a much complex domain, Atari, and learn features directly from raw input images instead of using hand-engineered features in a simpler domain. Learning directly from raw inputs is extremely challenging to an RL agent as it has to learn both the feature representations and the policy simultaneously.  

Leveraging human knowledge via learning from demonstration (LfD) is another effective way to pre-train an RL agent and has shown to be successful in robotics problems \citep{argall2009survey}. LfD has recently been applied widely in deep RL. \cite{christiano2017deep} uses human feedback to learn a reward function; \cite{hester2018learning} pre-train a Deep Q-network (DQN) \citep{mnih2015human} with human demonstrations by combining a large margin supervised loss with temporal difference loss, such that the agent closely imitates the demonstrator's policy at the beginning and later on learn to surpass the demonstrator. Our work similarly leverages a supervised loss as pre-training but differs in that we consider only the cross-entropy supervised loss as a feature learner, but not to imitate the policy. \cite{pohlen2018observe} builds upon \cite{hester2018learning} and proposes a reward-invariant update rule that uses the transformed Bellman (TB) operator in the DQN algorithm which better leverages expert demonstrations. In our work, we instead apply the TB operator in the A3C algorithm. The work of \cite{silver2016mastering} also trains human demonstrations in supervised learning then uses the supervised learner's network to initialize RL's policy network. However, their work uses a vast amount of expert demonstration data to train the supervised learner, while ours only uses a small amount of non-expert data. Our work is also the first to provide a comparative analysis on how pre-training with human data impacts learning in deep RL algorithms, as well as how our approach complements existing deep RL algorithms when (a small amount of) human demonstrations are available. 

\section{Background: Deep Reinforcement Learning}
\label{sec:background}
An RL problem is typically modeled using a Markov Decision Process (MDP) that is represented by a 5-tuple $\langle S, A, P, R, \gamma \rangle$. At each time step $t$, an RL agent receives some \emph{state} representation $S_t \in \mathcal{S}$ and explores an unknown environment by taking an \emph{action} $A_t \in \mathcal{A(}s\mathcal{)}$. A \emph{reward} $R_{t+1} \in \mathcal{R} \subset \mathbb{R}$ is given based on the action the agent took and the next state $S_{t+1}$ it reaches. The goal of an RL agent is to learn to maximize the expected return value $G_t = \sum_{k=0}^{\infty} \gamma^k R_{t+k+1}$ for each state at time $t$. The discount factor $\gamma \in [0,1]$ determines the relative importance of future and immediate rewards \citep{sutton2018reinforcement}.

The first successful deep RL method, deep Q-network (DQN), learns to play 49 Atari games directly from screen pixels by combining Q-learning with a deep convolutional neural network \citep{mnih2015human}. In classic Q-learning, an agent learns a state-action value function $Q^{\pi}(s, a) = \E_{s'}[r+\gamma \max_{a'}Q^{\pi} (s', a')|s, a]$, which is the expected discounted reward determined by performing action $a$ in state $s$ and thereafter following policy $\pi$ \citep{watkins1992q}. The optimal $Q^*$ can be deduced by following actions that have the maximum Q value, $Q^*(s,a) = argmax_{\pi}Q^{\pi}(s,a)$. Directly computing the $Q$ value is not feasible when the state space is large or continuous. The DQN algorithm uses a convolutional neural network as a function approximator to estimate $Q(s, a; \theta) \approx Q^*(s, a)$, where $\theta$ is the network's weight parameters. For each iteration $i$, DQN is trained to minimize
\[
L_{i}(\theta_i) = \E_{s, a, r, s'} \Big[(y - Q(s, a; \theta_i))^2 \Big]
\]
where $y = r + \gamma max_{a'}Q(s', a';\theta_{i}^-)$ is a \emph{target network} parameterized as $\theta_{i}^-$ that was generated from previous iterations. $\{s, a, r, s'\}$ are state-action samples drawn from an \emph{experience replay memory}, which is used to store the agent's experiences. At each time step, a batch of 32 samples (or \emph{minibatch}) is drawn from the experience replay memory to perform an update---this off-policy method could break the correlation between data (i.e., the sampled data is $i.i.d.$) which stabilizes the learning. All rewards are clipped to $[-1,1]$ to cope with different reward scale in games. The use of a target network, an experience replay memory, and the reward clipping are essential to stabilizing learning. The $\epsilon$-greedy policy is used by the agent to obtain sufficient exploration of the state space; for a probability of $\epsilon$, the agent selects a random action to explore.

\subsection{Asynchronous Advantage Actor-Critic (A3C)}

\begin{figure}[pt!]
    \centering
     \includegraphics[width=0.55\textwidth]{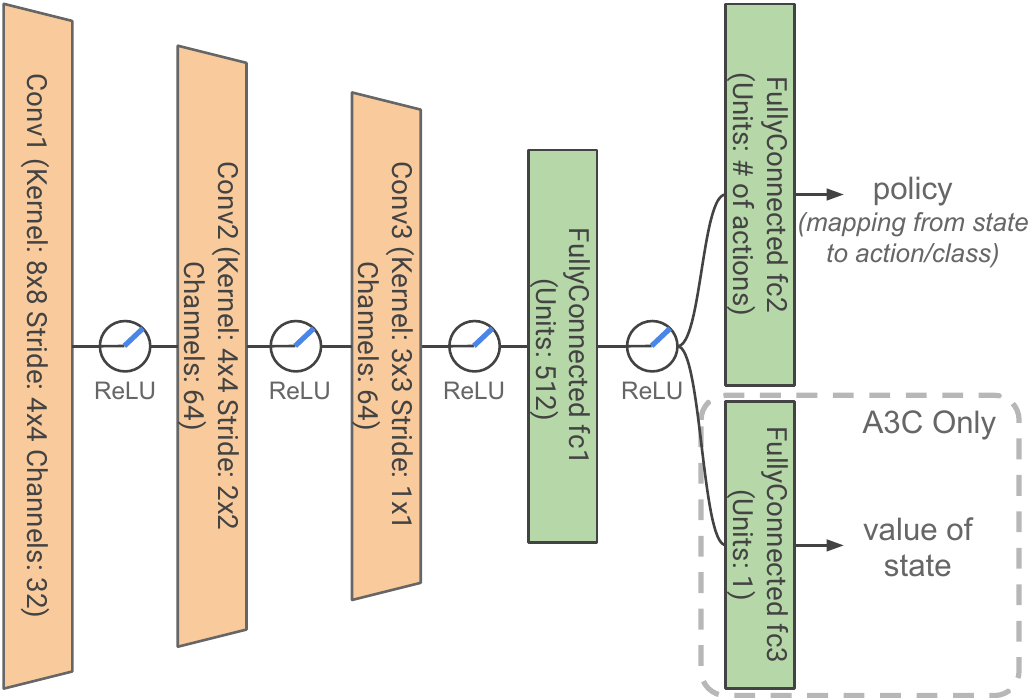} 
     \caption{Network architecture for each parallel actor in the A3C agent. We follow the same architecture as in \cite{mnih2015human} where there are three convolutional layers (\emph{conv1, conv2}, and \emph{conv3}), followed with two fully-connected layers (\emph{fc1} and \emph{fc2}), and a third fully-connected layer (\emph{fc3}) for learning the state value function as was done in \cite{mnih2016asynchronous}.} 
     \label{fig:network}
\end{figure}

The DQN algorithm suffers from two drawbacks: long training times and high computational requirements for memory and GPU resources. The A3C algorithm, in contrast, trains faster without the need of a GPU. In this work, we choose to use the A3C algorithm for all experiments.

A3C combines the actor-critic algorithm with deep RL. It differs from value-based algorithms where only a value function is learned. An actor-critic algorithm is policy-based and maintains both a policy function $\pi(a_t|s_t;\theta)$ and a value function $V^\pi(s_t;\theta_v)$. The policy function is called the \emph{actor}, which takes actions based on the current policy $\pi$. The value function is called the \emph{critic}, which serves as a baseline to evaluate the quality of the action using the state value $V^\pi(s_t;\theta_v)$. The network architecture of the A3C algorithm is shown in Figure \ref{fig:network}. There are three convolutional layers (\emph{conv1, conv2}, and \emph{conv3}), one fully connected layer of size $512$ (\emph{fc1}), followed by two branches of fully connected layer: \emph{fc2} is the policy function output layer which is of the same size as the number of actions and \emph{fc3} is the value function output layer of size $1$.

In A3C, $k$ actor-learners run in parallel with their own copies of the environment and parameters for the policy and value function, which enables exploration of different parts of the environment and therefore observations will not be correlated. Each actor-learner performs a parameter update every $t_{max}$ actions, or when a terminal state is reached---this is similar to using minibatch update as was done in DQN. Updates are synchronized to a master learner that maintains a central policy and value function, which will be the final policy upon the completion of training.

The policy network is directly parameterized and improved via policy-gradient \citep{sutton2018reinforcement}. To reduce the variance in policy gradient, an advantage function is used and calculated as $A(s_t, a_t;\theta,\theta_v) = Q^{(n)}(s_t, a_t;\theta,\theta_v) - V(s_t;\theta_{v})$. The $Q^{(n)}$ function is defined as

\begin{equation} \label{eq:q_function}
    Q^{(n)}(s_t, a_t;\theta,\theta_v) = \sum_{k=0}^{n-1} \gamma^{k}r_{t+k} + \gamma^{n}V(s_{t+n};\theta_v)
\end{equation}
where $n$ is upper-bounded by $t_{max}$. The loss function for the policy network is then defined as

\[L(\theta) = \nabla_\theta \log \pi(a_t|s_t;\theta)A(a_t, s_t;\theta,\theta_v) + \beta \nabla_\theta H(\pi(s_t;\theta))
\]
where $H$ is the entropy of policy $\pi$ that encourages exploration therefore helps prevent premature convergence to sub-optimal policies. The value network is updated using the loss function

\[L(\theta_v) = \nabla_{\theta_v} \Big[(Q^{(n)}(s_t, a_t;\theta,\theta_v) - V(s_t;\theta_{v}))^2 \Big]
\]

\subsection{Transformed Bellman Operator}
\label{sec:tb}
Reward clipping is introduced in DQN and is also used in A3C to cope with different reward scales among Atari games \citep{mnih2015human,mnih2016asynchronous}. However, this is problematic because the RL agent will not be able to distinguish between states with high rewards versus those with low rewards, resulting in learning a suboptimal policy. The poor performance in some Atari games has been attributed to reward clipping \citep{hester2018learning}.

%\MET{Define $\varepsilon$?}\gabe{modified}
In this work, we apply the transformed Bellman operator \citep{pohlen2018observe} to the A3C algorithm to overcome the problem of reward clipping. We use the raw rewards instead of clipping them to the scale of $[-1,1]$. A function $h:\mathbb{R} \rightarrow \mathbb{R}$ is used to reduce the scale of $Q^{(n)}(s_t, a_t;\theta,\theta_v)$ (Equation~\ref{eq:q_function}) and is transformed as

\begin{equation} \label{eq:tb_q_function}
    Q^{(n)}(s_t, a_t;\theta,\theta_v) = \sum_{k=0}^{n-1} h\left( \gamma^{k}r_{t+k} + \gamma^{n}h^{-1}\left( V\left(s_{t+n};\theta_v\right) \right) \right)
\end{equation}

\begin{equation} \label{eq:h_function}
    h : z \mapsto sign(z)\left(\sqrt{|z| + 1} - 1\right) + \varepsilon z
\end{equation}

\begin{equation} \label{eq:h_inv_function}
    h^{-1} : x \mapsto sign(x) \left(\left(\frac{\sqrt{1+4\varepsilon(|x|+1+\varepsilon)} - 1}{2\varepsilon}\right)^2 - 1\right)
\end{equation}

\noindent where $\varepsilon z$ is for regularization that ensures $h^{-1}$ is Lipschitz continuous and a closed form inverse.

\section{Supervised Pre-Training for Deep RL}
\label{sec:pretrain}

Deep reinforcement learning can be divided into two sub-tasks: feature learning and policy learning. Although deep RL in itself has succeeded in learning both tasks simultaneously, it still suffers from long training time and slow learning. We believe that by addressing feature learning, we can jumpstart the performance in an RL agent since it will be able to focus more on policy learning, which in turn speeds up the entire learning process. 

In this article, we propose to use supervised pre-training on human demonstration data as a way to address feature learning. We train a multiclass-classification network over a set of non-expert human demonstrations, where actions demonstrated by the human were used as the ground truth labels for a given input game frame. The network uses the same architecture as in A3C (shown in Figure~\ref{fig:network}) where the fc2 layer is used as the classification output layer. Note that we exclude the fc3 layer for the classification task. The network classifier minimizes a softmax cross entropy loss using the RMSProp \citep{tieleman2012lecture} optimizer with a set of hyperparameters shown in Table~\ref{table:hyperparameters}. We also use gradient clipping and L2 regularization for more stable training.

However, we identify two problems when using non-expert human demonstration for pre-training. First, we assume the actions provided by the human are the correct labels---the low quality of our data shown in Table~\ref{table:human_demo} indicates that we are pre-training with noisy data. In this work, we empirically study if the noise in the data would still allow us to learn important features. Second, the collected human data is highly imbalanced. For example, in the game of Breakout, after hitting the ball, the human usually do nothing until the ball bounces and starts falling back to the paddle, which results in most collected actions being the ``NOOP'' action. In some games, the human demonstrator tends to use the simpler actions like ``LEFT'' instead of the compound actions like ``LEFTFIRE''. The class imbalance problem plagues all six %\gabe{modified}\MET{most? many?} 
games used in our experiments. To cope with this, we use proportional sampling. During minibatch sampling, we randomly sample over all available actions based on their proportion to the entire demonstration data; doing so ensures that each minibatch includes a relatively balanced set of classes. We pre-train a classification model for each Atari game for 750,000 training iterations.

After pre-training, the learned weights and biases from the classifier are then used to initialize the A3C's network (instead of random initialization). Note that when using all layers' parameters from the pre-trained model (including the output \emph{fc2} layer), normalizing the output layer's weights is necessary to achieve a positive result; we empirically observe that the values of the output layer tend to explode without normalization. To normalize the output layer, we keep track of the maximum value of the output layer during training, which is then used as the divisor to all weights and biases. We refer to our pre-training method as the \emph{pre-trained model for A3C (PMfA3C)}. We also apply pre-training to the transformed Bellman operator variant of the A3C algorithm, and we refer to it as \emph{PMfA3C-TB}.

\section{Experimental Design}
\label{sec:experiment}
We evaluate our approach in six Atari games: Asterix, Breakout, MsPacman, NameThisGame, Pong, and SpaceInvaders. We use the deterministic version four of the Atari 2600 environment from OpenAI Gym \citep{openaigym}. We follow the Atari settings described in the DQN algorithm and OpenAI baselines Atari wrapper~\citep{mnih2015human, baselines}. Here are the Atari settings used:
\begin{itemize}
    \item At the beginning of a game, the agent executes a random $x$ ($0 \leq x \leq 30$) number of ``NOOP'' actions.
    \item Take an action for games (e.g., Breakout) that remains static unless pressing ``FIRE.''
    \item Apply max pooling over the game frames to remove flickering effects of the game.
    \item Consider loss of life as the end of an episode or as a terminal state, but only do a hard-reset on the game environment (i.e., reset back to the initial game state) when losing all lives.%, and
    \item Use a frame skip of four, meaning that an action is repeated for four frames before a new action is selected.
\end{itemize}

The network architecture for A3C is shown in Figure~\ref{fig:network}. The four most recent game frames are used as input to the network, each frame is converted to grayscale and resized to $84 \times 84$ without cropping. We use the same set of hyperparameters for all games (except for Pong). We summarize the hyperparameter values in Table~\ref{table:hyperparameters}. Gradient clipping is also used in A3C. For all experiments, we train a total of $50$ million steps, distributed over $16$ parallel A3C actors. Each step consists four game frames (since we use frame skip of four) thus all experiments run a total of 200 million game frames.  

\begin{table}[h!]
    \caption{All games use the same set of hyperparameters except for Pong, where we found setting RMSProp epsilon to $1\times10^{-4}$ gives a much more stable learning.}
    \label{table:hyperparameters}
    \centering
    \adjustbox{max width=\textwidth}{
        \begin{tabular}{r|r}
        \textbf{Parameter}        & \textbf{Value} \\ \hline \hline
        RMSProp learning rate  & $7 \times 10^{-4}$  \\ \hline
        RMSProp epsilon        & $1 \times 10^{-5}$  \\ \hline
        RMSProp decay          & 0.99      \\ \hline 
        RMSProp momentum       & 0      \\ \hline 
        Maximum gradient norm  & 0.5     \\ \hline \hline
        \multicolumn{2}{c}{\textbf{Parameters unique to supervised pre-training}} \\ \hline \hline
        Number of mini-batch updates & 750,000     \\ \hline
        Batch size             & 32     \\ \hline
        L2 regularization weight & $1 \times 10^{-4}$      \\ \hline \hline
        \multicolumn{2}{c}{\textbf{Parameters unique to A3C only}} \\ \hline \hline
        $k$ parallel actors  & 16      \\ \hline
        $t_{max}$  & 20          \\ \hline
        transformed Bellman operator $\varepsilon$ & $10^{-2}$ \\ \hline \hline
        \end{tabular}
    }
\end{table}

\subsection{Collection of Human Demonstration}
We use the keyboard interface from OpenAI Gym \citep{openaigym} to enable interactions between the human demonstrator and the Atari environment. For each game, the demonstrator is provided with game rules and a set of valid actions with their corresponding keyboard keys. The frame skip is set to one to provide smoother game transitions during human plays (whereas we reset it to four during agent training). To simulate frame skipping during the demonstration, we collect every fourth frame of the game. At each collection step, we save: 1) the game image (i.e., the state), 2) the action taken by the demonstrator, 3) the reward received, and 4) if the current state is a terminal state. For each episode, we allow a maximum of 20 minutes of playing time for the human demonstrator. The demonstration ends when the game reaches the time limit or when the game ends---whichever comes first. Table~\ref{table:human_demo} provides a breakdown of the demonstration size and quality for all games. 

\subsection{Evaluation Procedures}
For all experiments, we perform policy evaluation on the RL agent at every one million training steps. We get the average testing score over 125,000 testing steps and report the average over four trials. We report the highest average reward of the RL agent and also measure the learning speed improvement using three metrics adapted from \cite{taylor2009transfer}: 
\begin{enumerate}
    \item \emph{Best reward}: the highest average reward attained by the agent from over four trials.
    \item \emph{Final performance}: the final learned performance of the agent. We use the reward obtained at step $50$ million as the value for the final performance. 
    \item \emph{Total reward}: the total reward accumulated (i.e., the area under the learning curve (AUC)) by the agent. We approximate the AUC using the \emph{trapezoidal rule}: $AUC \approx \sum^{T}_{t=1} \frac{f(x_{t-1}) + f(x_t)}{2} \Delta x_t$, where $f(x_t)$ is the reward value at time $t$ and $\Delta x_t = x_t - x_{t-1} = 10^6$ is the evaluation frequency. Note that for readability, we scale down all calculation results by one million and consider $\Delta x_t = 1$, this does not affect the comparison results.
    \item \emph{Reward improvement}: the ratio of the total reward improvement of the pre-trained agent compared to the baseline agent. We calculate it as $\% Ratio = \frac{AUC_{\text{pre-trained}} - AUC_{\text{baseline}}}{AUC_{\text{baseline}}} \times 100$
\end{enumerate}

\begin{table}[pt!]
    \caption{Human demonstration size and quality. The data is of a small amount and the demonstrator is a non-expert, compared to the best human demonstration score in the state-of-the-art Ape-X DQfD algorithm \citep{pohlen2018observe} (Ape-X DQfD did not collect human demonstration for SpaceInvaders).}
    \label{table:human_demo}
    \centering
    \adjustbox{max width=\textwidth}{
        \begin{tabular}{r|r|r|r|r|r}
        Game          & Worst score & Best score   & Best score  & \# of states & \# of episodes \\
                      &             &              &  (Ape-X DQfD) &  &  \\ \hline \hline
        Asterix       & 6250        & 14300        & $\mathbf{18100}$  & 12870       &  5  \\ \hline
        Breakout      & 26          & 59           & $\mathbf{79}$     & 10190       &  10 \\ \hline
        % Freeway     & 30          & 31           & $\mathbf{32}$     & 10176       &  5  \\ \hline
        % Gopher      & 1420        & 5800         & $\mathbf{22520}$  & 16847       &  8  \\ \hline
        MsPacman      & 4020        & 18241        & $\mathbf{55021}$  & 14504       &  8  \\ \hline
        NameThisGame  & 2510        & 4840         & $\mathbf{19380}$  & 17113       &  4  \\ \hline
        Pong          & -13         & $\mathbf{5}$ & 0      & 21674       &  6  \\ \hline
        SpaceInvaders & 545         & 1840         & -      & 16807       &  8  \\ \hline \hline
        \end{tabular}
    }
\end{table}

\section{Results}
\label{sec:results}
\begin{figure}[pt!]
  \centering
    \begin{subfigure}[h]{\textwidth}
        \includegraphics[width=0.33\textwidth]{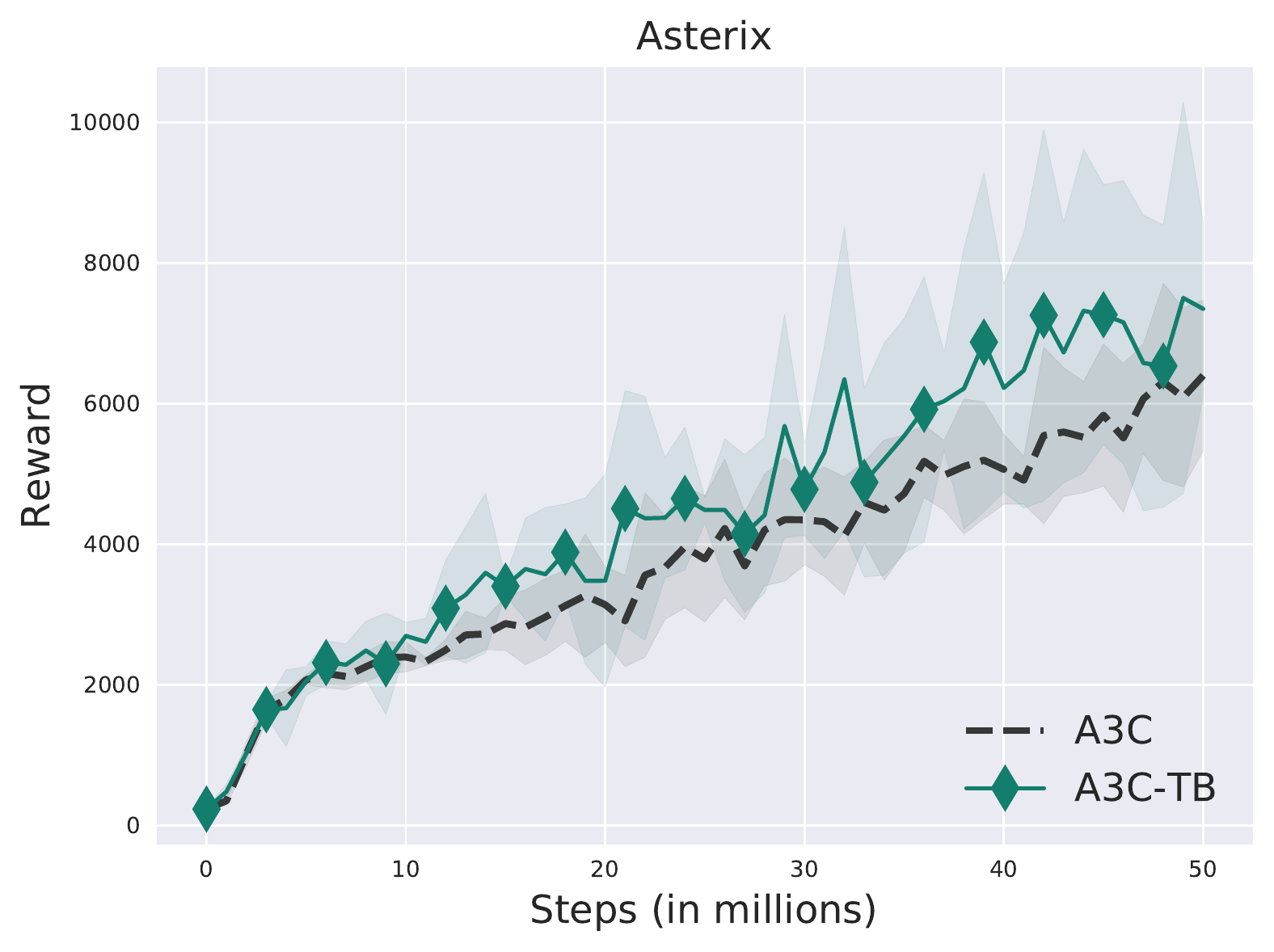}
        \includegraphics[width=0.33\textwidth]{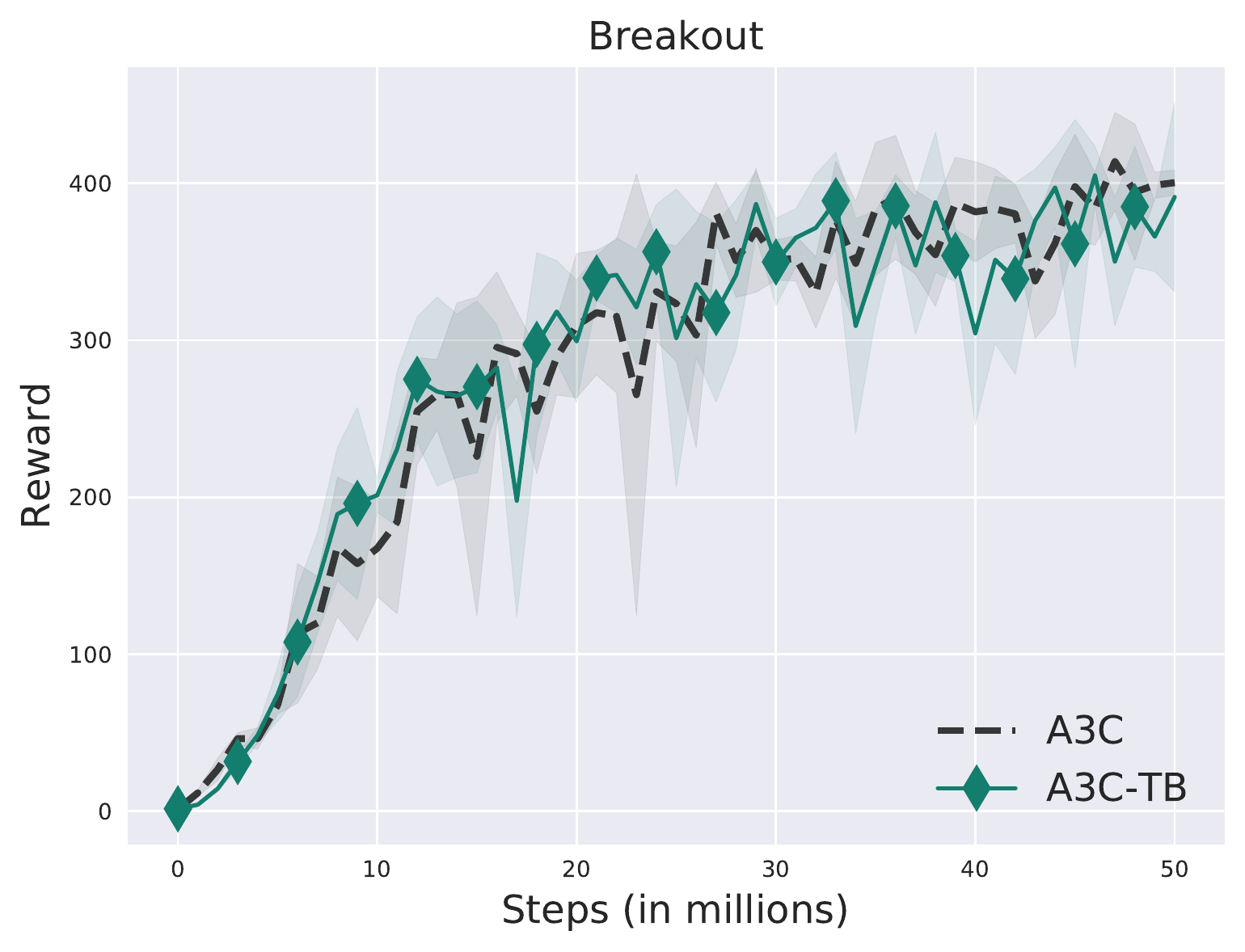}
        \includegraphics[width=0.33\textwidth]{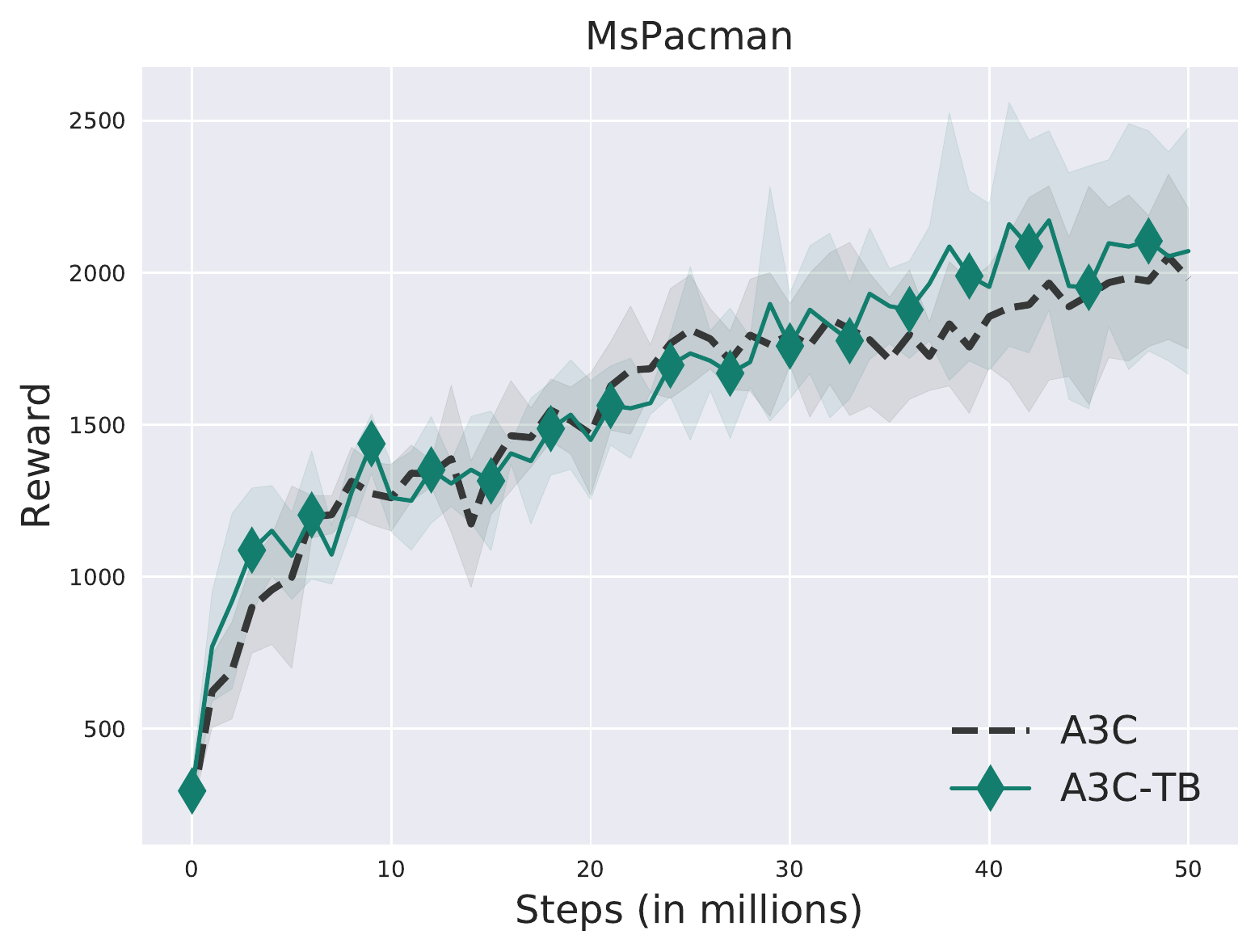}
    \end{subfigure}

    \begin{subfigure}[h]{\textwidth}
        \includegraphics[width=0.33\textwidth]{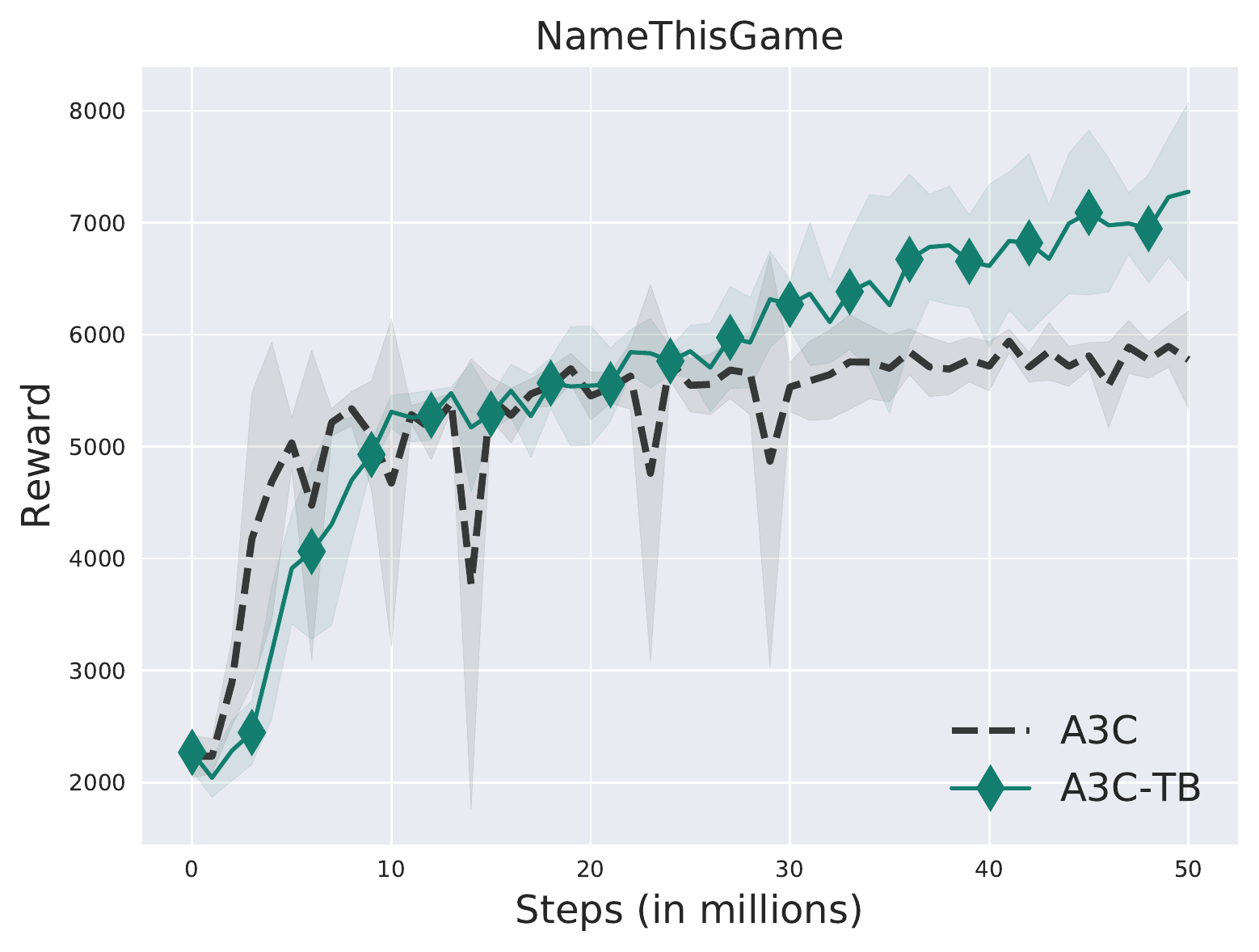}
        \includegraphics[width=0.33\textwidth]{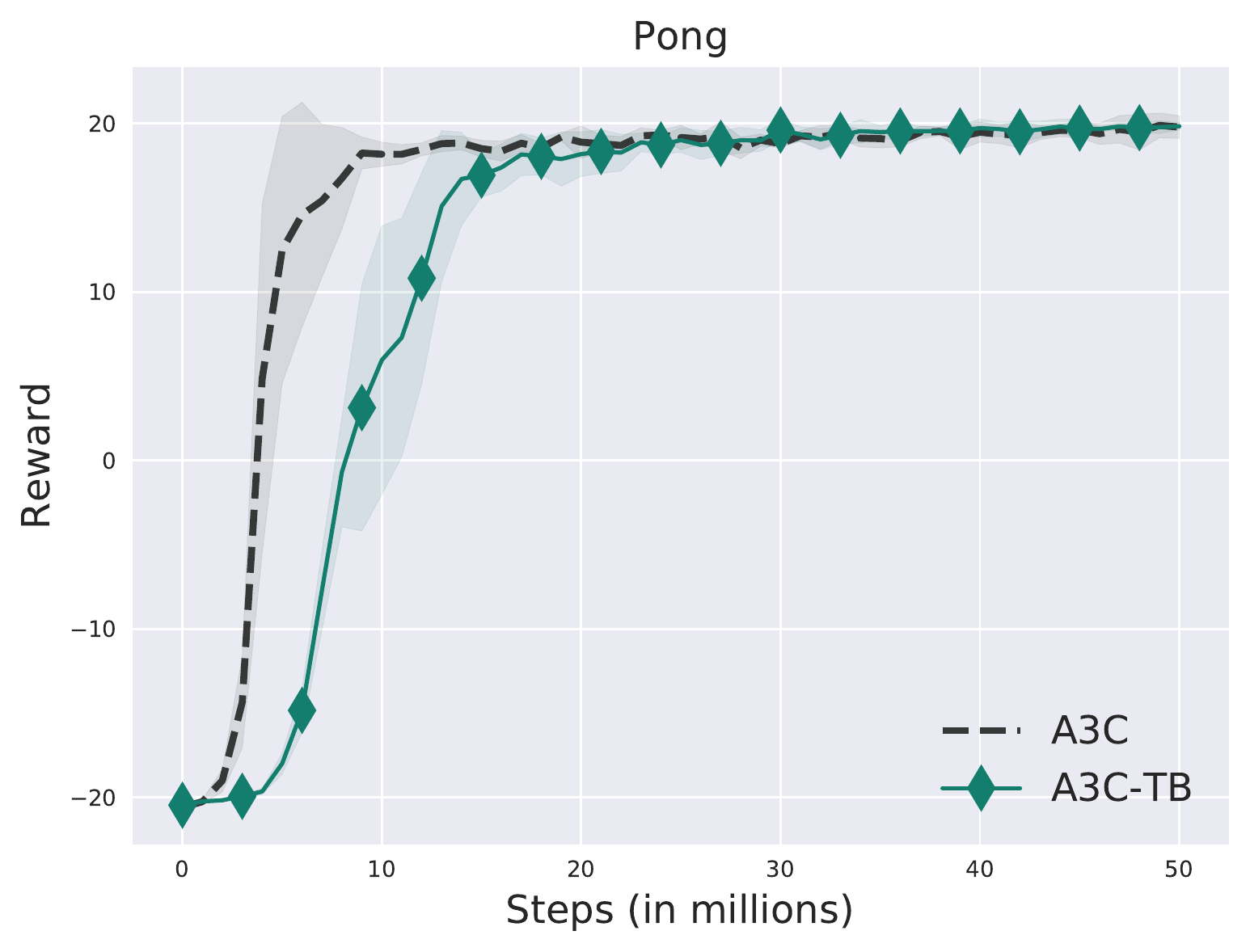}
        \includegraphics[width=0.33\textwidth]{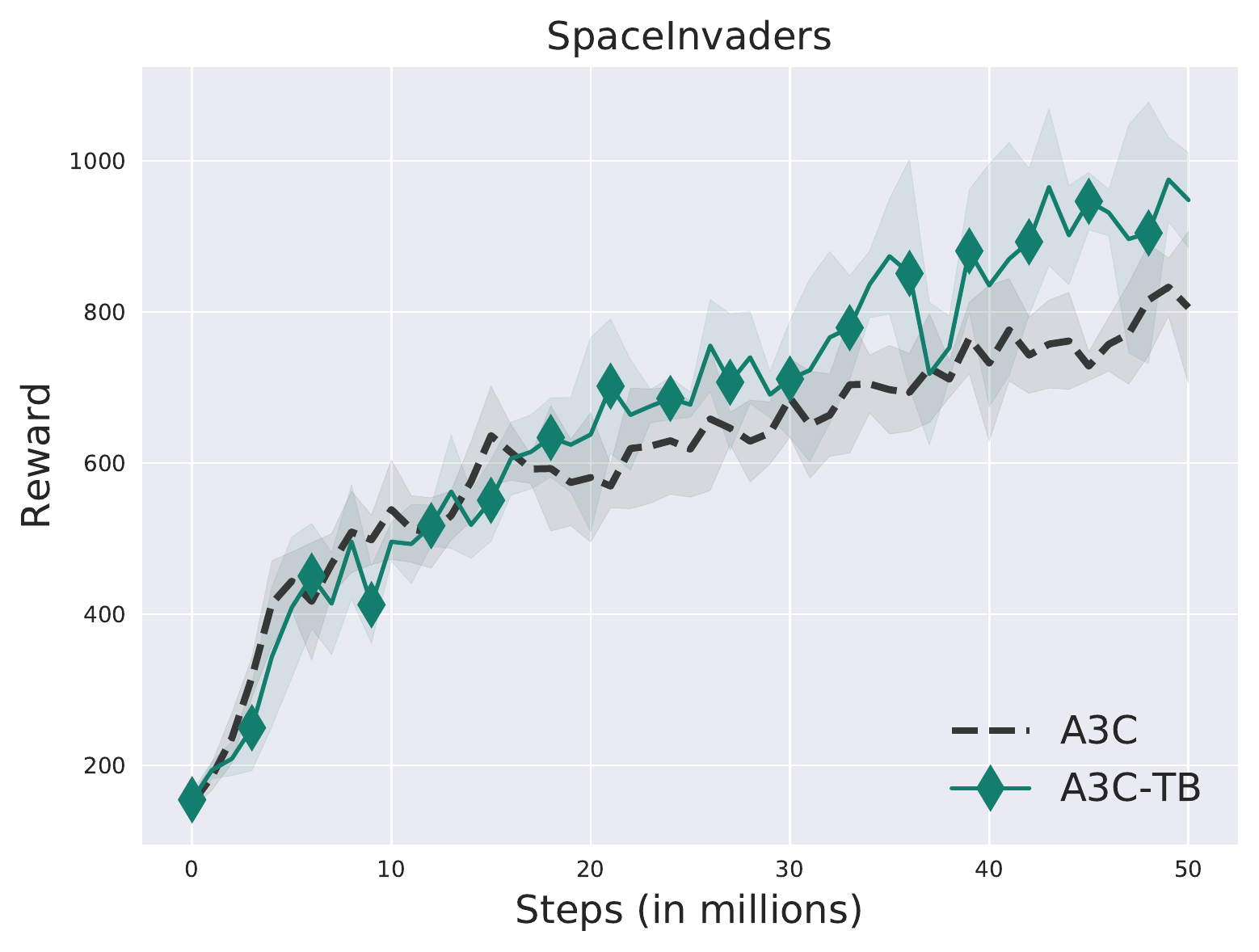}
    \end{subfigure}
    
    \caption{Performance of the baseline A3C and the transformed Bellman operator variant A3C (A3C-TB). The x-axis is the total number of training steps (among all $16$ actors), where each step consists of four game frames (we use frame skip of four). The y-axis is the average testing score over four trials where the shaded regions correspond to the standard deviation.}
    \label{fig:a3c_base_result}
\end{figure}

First, we present the performance of the baseline A3C and the transformed Bellman operator variant A3C (A3C-TB). Figure~\ref{fig:a3c_base_result} shows that A3C-TB outperforms the baseline A3C in five out of the six games. Although Pong in A3C-TB has a low performance,\footnote{In the game of Pong, its rewards $r \in \{-1, 0, 1\}$ are already at the same reward scale as applying the reward clipping in A3C at $r \in [-1,1]$. Thus, reward clipping has no effect in Pong's performance. However, applying the transformed Bellman operator scales down Pong's reward values, which slows down the reward propagation. We believe this is the reason why Pong's performance is negatively affected in A3C-TB.} we still consider the results as consistent with the findings in \cite{hester2018learning} that using reward clipping leads to an agent learning a suboptimal policy. As we discussed in Section \ref{sec:tb}, A3C-TB enables using raw reward signals such that the RL agent can distinguish between low and high rewarding states, which leads to better policy learning. This is even more important to address when learning from demonstrations for games that have distinct reward signals. For example in the game of MsPacman, human tends to take actions that move towards states with high rewards (e.g., eating an edible ghost is more rewarding than eating the dots). However, in a baseline A3C agent where the rewards are clipped, it sees all rewards as equal and might not be able to leverage the human knowledge of ``eating an edible ghost.'' 

Next, we present and discuss results for our pre-training approaches, \emph{PMfA3C} and \emph{PMfA3C-TB}. Note that we do not compare our results to recent algorithms in \cite{hester2018learning} and \cite{pohlen2018observe} since our training steps are much shorter and are not directly comparable to those that were trained on large-scale.

%\MET{Does the x-axis show the number of frames or the number of actions taken? Won't these be different by a factor of 4? (I'm not sure what a `step' means.)}
%\yunshu{changed caption. 1 step = 1 action, 1 action is repeated 4 times (i.e., frames)}

\begin{figure}[pt!]
  \centering
    \begin{subfigure}[h]{\textwidth}
        \includegraphics[width=0.33\textwidth]{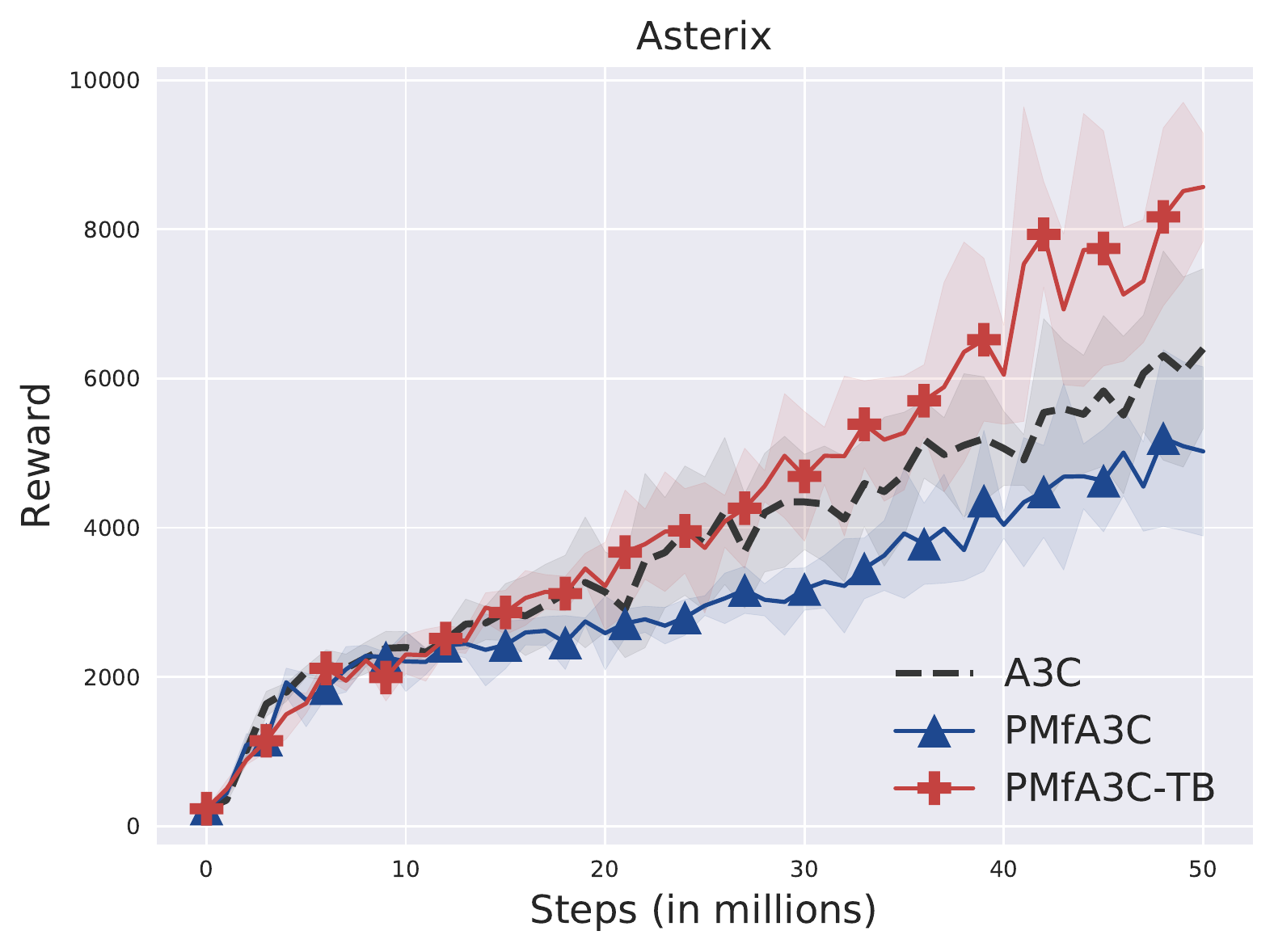}
        \includegraphics[width=0.33\textwidth]{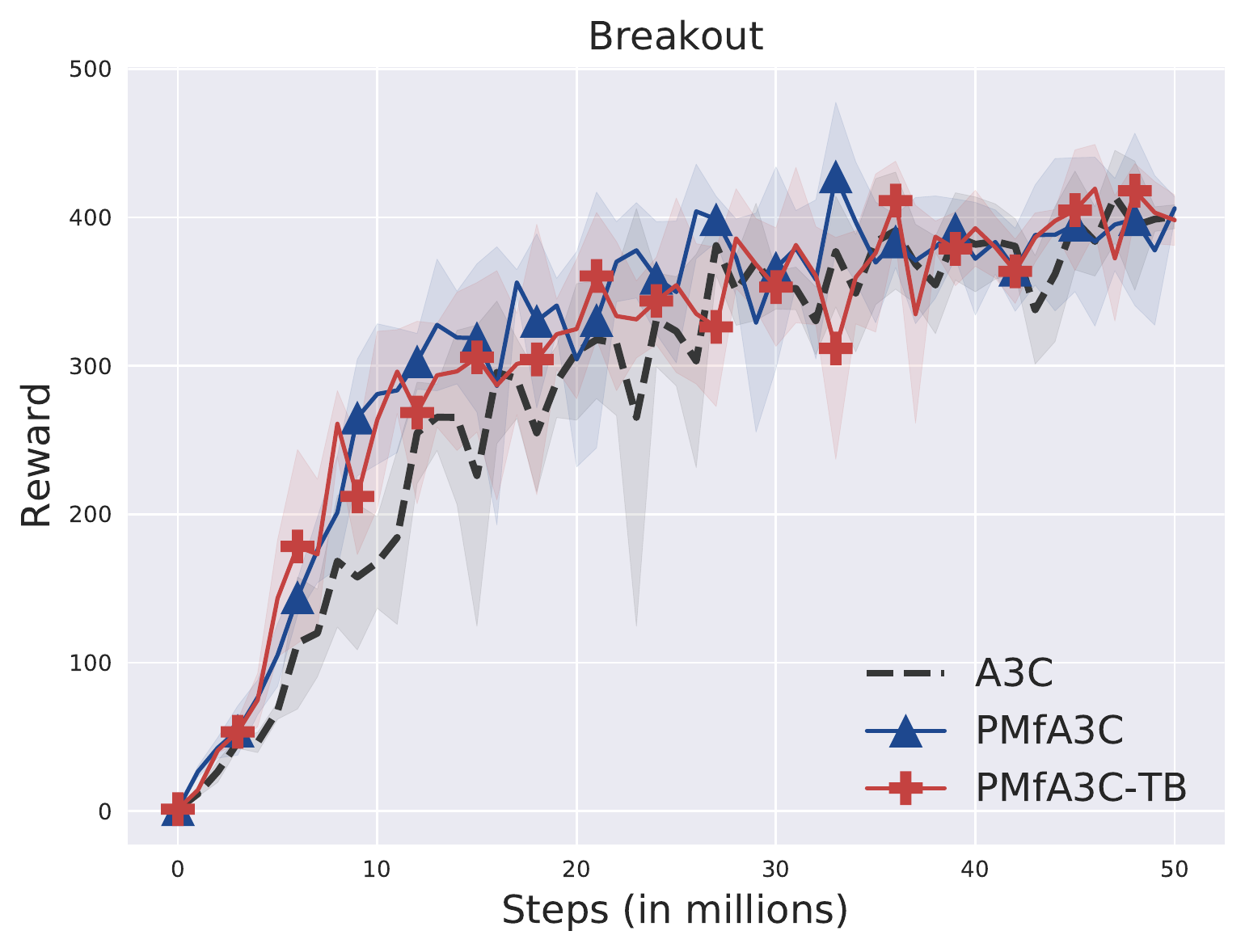}
        \includegraphics[width=0.33\textwidth]{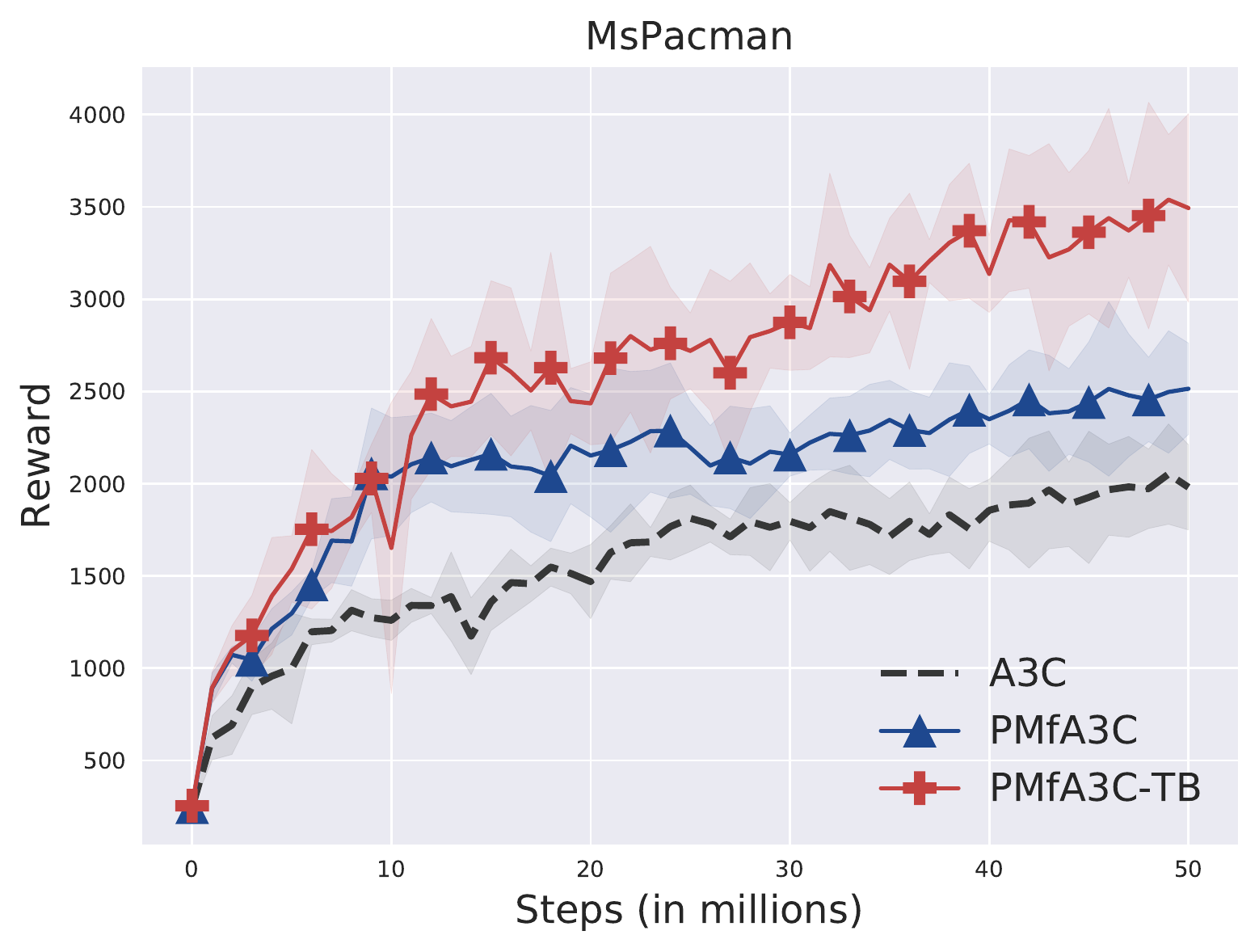}
    \end{subfigure}

    \begin{subfigure}[h]{\textwidth}
        \includegraphics[width=0.33\textwidth]{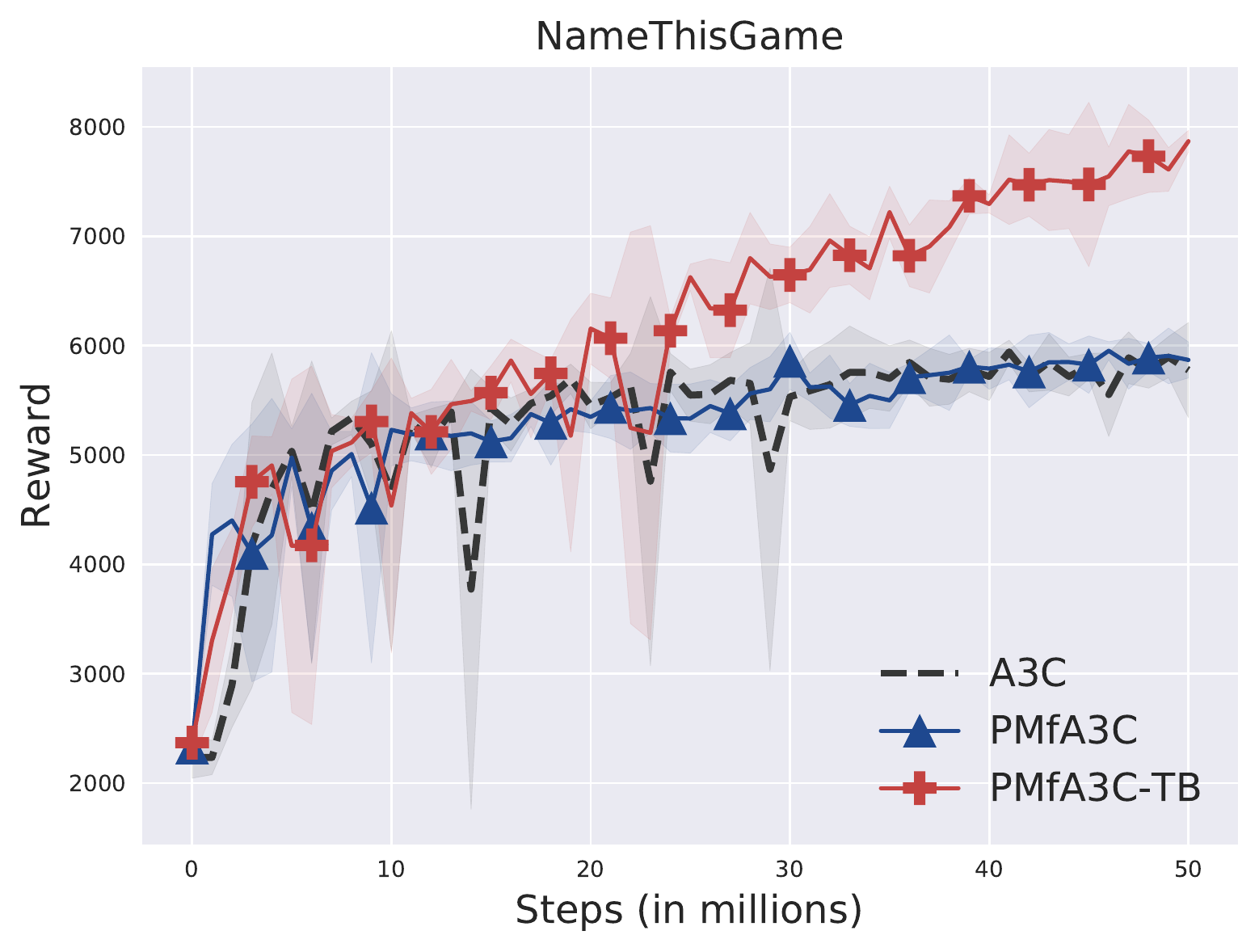}
        \includegraphics[width=0.33\textwidth]{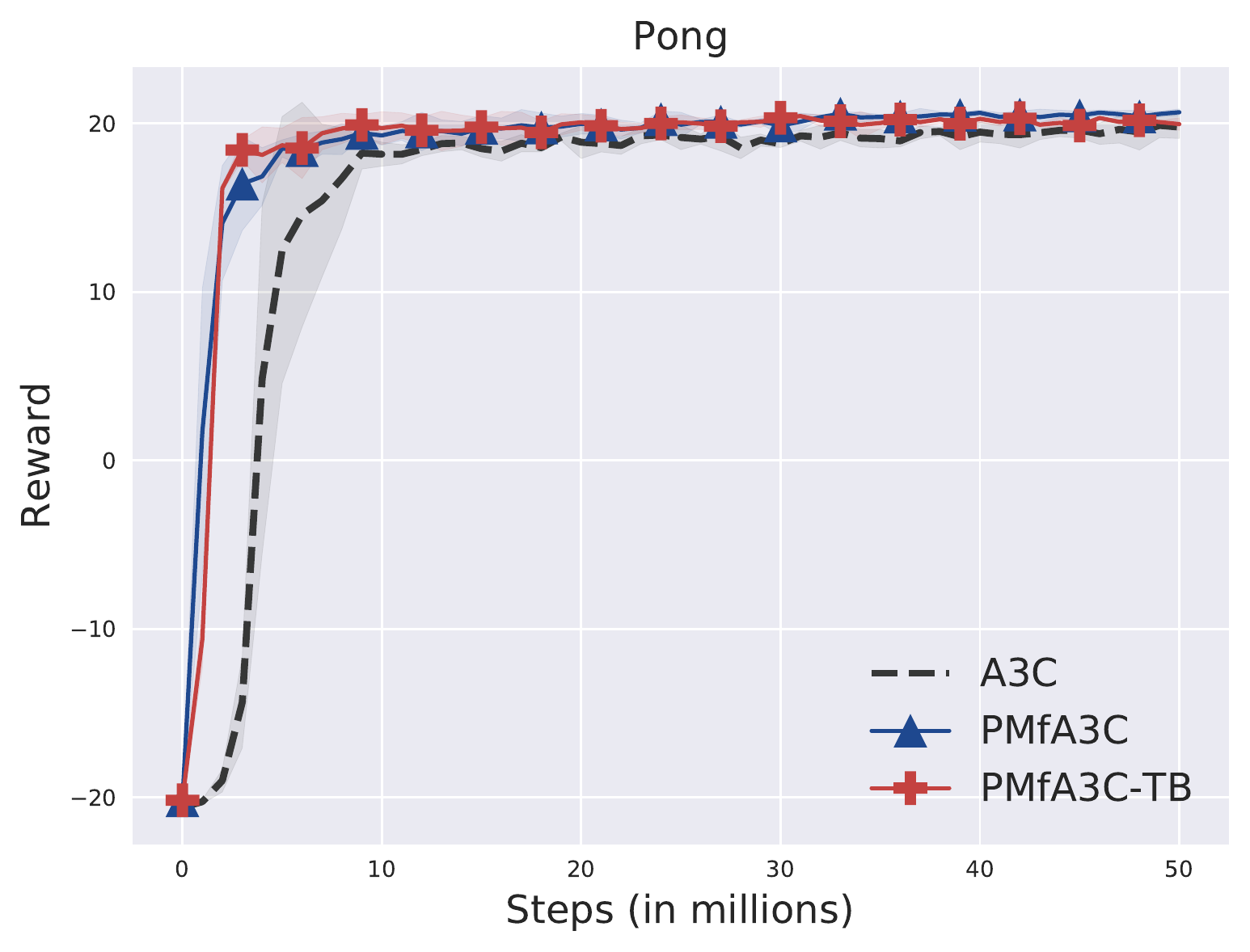}
        \includegraphics[width=0.33\textwidth]{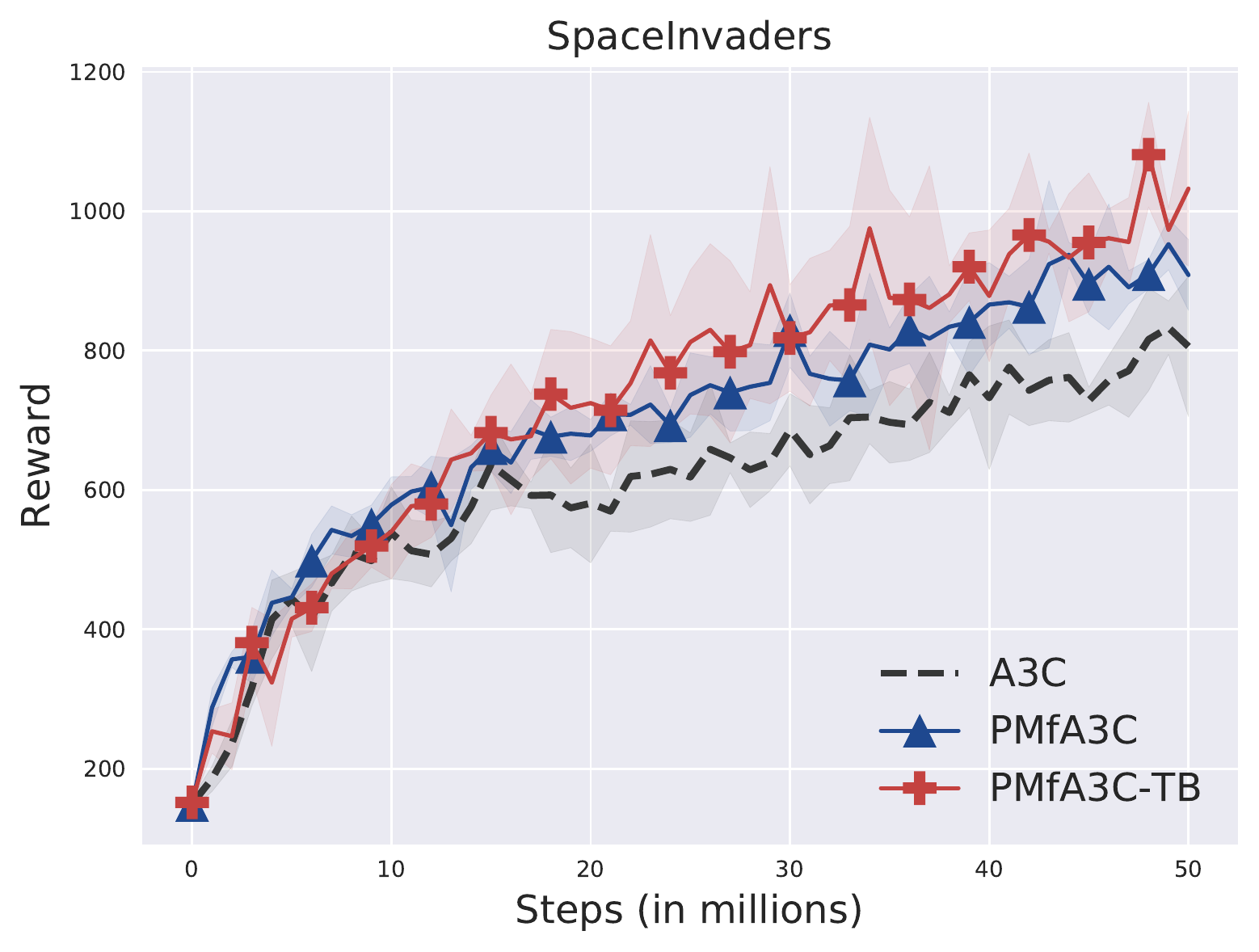}
    \end{subfigure}
    \caption{Performance of baseline and pre-training using A3C. The x-axis is the total number of training steps (among all $16$ actors), where each step consists of four game frames (we use frame skip of four). The y-axis is the average testing score over four trials where the shaded regions correspond to the standard deviation.}
    \label{fig:a3c_result}
\end{figure}

\subsection{Pre-Training Methods}
\label{sec:pretrain-results}

We apply pre-training methods as described in Section \ref{sec:pretrain} to both PMfA3C and PMfA3C-TB. A multiclass-classification network is pre-trained using the human demonstration dataset. All weights and biases from the classifier are used to initialize A3C's network layers, conv1, conv2, conv3, fc1, and fc2; the fc3 layer is initialized randomly. Figure~\ref{fig:a3c_result} shows the learning curve of both PMfA3C and PMfA3C-TB. Compared to the baseline A3C, PMfA3C outperforms in three out of six games; PMfA3C-TB shows a much stronger performance and exceeds the baseline in five out of six games. 

Table~\ref{table:performance} shows the quantitative performance improvements of our pre-training methods over the baseline. 
PMfA3C and PMfA3C-TB achieve the best performance in the game of MsPacman among all experiments with a remarkable total reward improvement of $67.36\%$. This verifies the importance of being able to learn from raw reward signals (instead of clipped rewards) in games with various reward scales.
% \MET{Doesn't it show that raw rewards + LfD is better than clipped rewards + no LfD? It seems like we'd want to have 1) clipped rewards + no LfD, 2) clipped rewards + LfD, 3) raw rewards + no LfD, 4) raw rewards + LfD. I'm not sure it's worth running these experiments, but just be careful about what you claim from these results.} 
SpaceInvaders and Pong also show good reward improvement ratios in both pre-training methods compared to the baseline. While the performance increase in Breakout might not seem obvious visually from the learning curves, their total rewards indicate a $10.64\%$ and an $8.06\%$ improvement for PMfA3C and PMfA3C-TB respectively. Contrarily, we note that PMfA3C did not help in NameThisGame and Asterix. The former shows comparable results as the baseline A3C with a negligible improvement of $0.91\%$. When using PMfA3C for Asterix, it shows the worst performance among all experiments with an $18.01\%$ performance drop compared to the baseline. However, we point out that when applying the TB operator, A3C-TB and PMfA3C-TB, Asterix outperforms the baseline by $21.08\%$ and $15.38\%$ respectively. 

In summary, our experiment results show that: 1) using the TB operator is important for games with various reward scales and 2) pre-training is beneficial for training in the RL agent.

\begin{table}[pt!]
\caption{Quantitative evaluation of pre-training methods using four metrics: best reward, final performance, total reward, and reward improvement.}
\label{table:performance}
\centering
\adjustbox{max width=\textwidth}{
\begin{tabular}{r|r|r|r|r|r}
\multicolumn{2}{c|}{\multirow{2}{*}{Game}} & \multirow{2}{*}{Best reward} & \multirow{2}{*}{Final performance} & \multirow{2}{*}{Total reward}  & Reward   \\ 
\multicolumn{2}{c|}{}  &    & & & improvement   \\ \hline \hline
\multirow{4}{*}{Asterix}       & A3C         & $6398.44$          & $6398.43 \pm 1072.22$ & $187702.21 \pm 26249.69$          & -                   \\ \cline{2-6}
                               & A3C-TB      & $7500.58$          & $7345.85 \pm 1258.83$ & $\mathbf{223942.78 \pm 47077.93}$ & $\mathbf{21.08\%}$  \\ \cline{2-6}
                               & PMfA3C      & $5201.08$          & $5022.76 \pm 1134.26$ & $153889.07 \pm 15016.30$          & $-18.01\%$          \\ \cline{2-6}
                               & PMfA3C-TB   & $\mathbf{8566.49}$ & $\mathbf{8566.49 \pm 724.55}$ & $216571.05 \pm 18629.16$  & $15.38\%$           \\ \hline
\multirow{4}{*}{Breakout}      & A3C         & $413.88$           & $400.40 \pm 8.04$     & $14198.48 \pm 606.64$             & -                   \\ \cline{2-6}
                               & A3C-TB      & $405.03$           & $391.23 \pm 60.39$    & $14197.93 \pm 181.07$             & $-0.29\%$           \\ \cline{2-6}
                               & PMfA3C      & $\mathbf{427.56}$  & $\mathbf{406.03 \pm 7.83}$  & $\mathbf{15709.87 \pm 171.10}$ & $\mathbf{10.64\%}$  \\ \cline{2-6}
                               & PMfA3C-TB   & $419.28$           & $398.13 \pm 16.98$    & $15343.39 \pm 472.31$             & $8.06\%$            \\ \hline
\multirow{4}{*}{MsPacman}      & A3C         & $2052.07$          & $1980.46 \pm 231.12$  & $78419.18 \pm 6027.52$            & -                   \\ \cline{2-6}
                               & A3C-TB      & $2172.01$          & $2071.19 \pm 405.62$  & $80959.67 \pm 6773.66$            & $3.24\%$            \\ \cline{2-6}
                               & PMfA3C      & $2514.61$          & $2514.61 \pm 247.60$  & $103950.65 \pm 8529.05$           & $32.56\%$           \\ \cline{2-6}
                               & PMfA3C-TB   & $\mathbf{3539.0}$  & $\mathbf{3493.90 \pm 510.68}$ & $\mathbf{131239.08 \pm 9851.32}$  & $\mathbf{67.36\%}$  \\ \hline
\multirow{4}{*}{NameThisGame}  & A3C         & $5942.29$          & $5775.73 \pm 435.45$  & $264140.63 \pm 9830.70$           & -                   \\ \cline{2-6}
                               & A3C-TB      & $7276.32$          & $7276.32 \pm 799.27$  & $282540.94 \pm 14059.33$          & $6.97\%$            \\ \cline{2-6}
                               & PMfA3C      & $5952.68$          & $5868.82 \pm 164.64$  & $266544.47 \pm 5827.28$           & $0.91\%$            \\ \cline{2-6}
                               & PMfA3C-TB   & $\mathbf{7869.28}$ & $\mathbf{7869.28 \pm 99.12}$ & $\mathbf{306014.77 \pm 4438.13}$  & $\mathbf{15.85\%}$  \\ \hline
\multirow{4}{*}{Pong}          & A3C         & $19.89$            & $19.79 \pm 0.67$      & $791.71 \pm 32.08$                & -                   \\ \cline{2-6}
                               & A3C-TB      & $19.84$            & $19.84 \pm 0.23$      & $604.60 \pm 46.49$                &  $-23.63\%$         \\ \cline{2-6}
                               & PMfA3C      & $\mathbf{20.67}$   & $\mathbf{20.67 \pm 0.20}$   & $\mathbf{948.38 \pm 8.21}$        & $\mathbf{19.79\%}$  \\ \cline{2-6}
                               & PMfA3C-TB   & $20.44$            & $19.96 \pm 0.37$      & $937.61 \pm 19.44$                & $18.43\%$           \\ \hline
\multirow{4}{*}{SpaceInvaders} & A3C         & $832.98$           & $805.96 \pm 123.04$   & $30533.94 \pm 1413.44$            & -                   \\ \cline{2-6}
                               & A3C-TB      & $975.27$           & $948.26 \pm 70.67$    &  $33293.51 \pm 1506.37$           & $9.04\%$            \\ \cline{2-6}
                               & PMfA3C      & $952.71$           & $908.50 \pm 50.73$    & $35163.27 \pm 697.12$             & $15.16\%$           \\ \cline{2-6}
                               & PMfA3C-TB   & $\mathbf{1081.21}$ & $\mathbf{1032.48 \pm 111.07}$ & $\mathbf{36902.61 \pm 2529.46}$ & $\mathbf{20.86\%}$  \\ \hline \hline
\end{tabular}}
\end{table}

\subsection{Ablation Studies}
In reusing networks for image classification, \cite{yosinski2014transferable} pointed out that the lower layers of a network tend to learn more general features while upper layers tend to learn more specific features towards the task. This is what inspired us to use all layers from the pre-trained network since both networks are trained on data collected from the same game.

\begin{figure}[pt!]
  \centering
    \begin{subfigure}[h]{\textwidth}
        \includegraphics[width=0.33\textwidth]{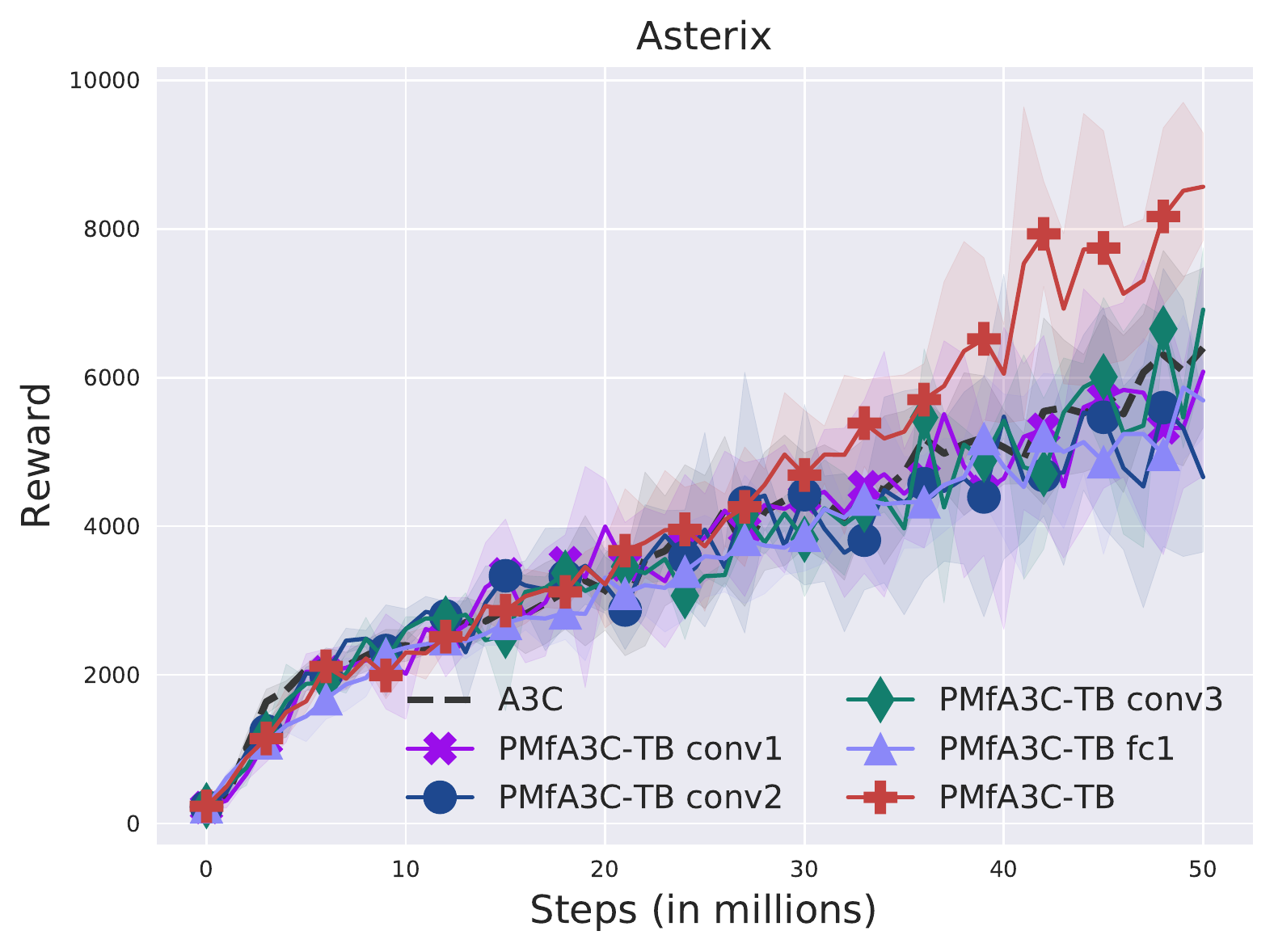}
        \includegraphics[width=0.33\textwidth]{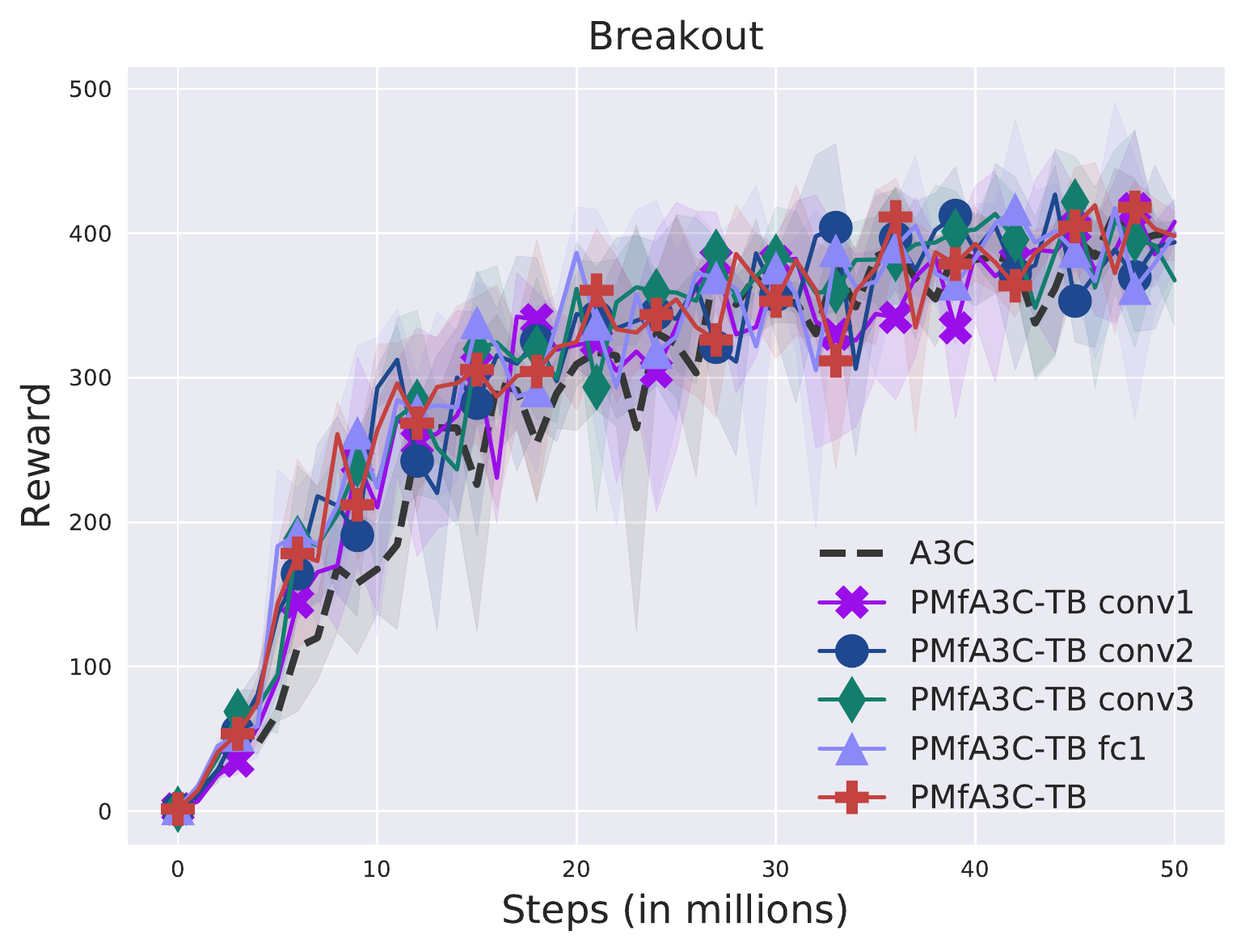}
        \includegraphics[width=0.33\textwidth]{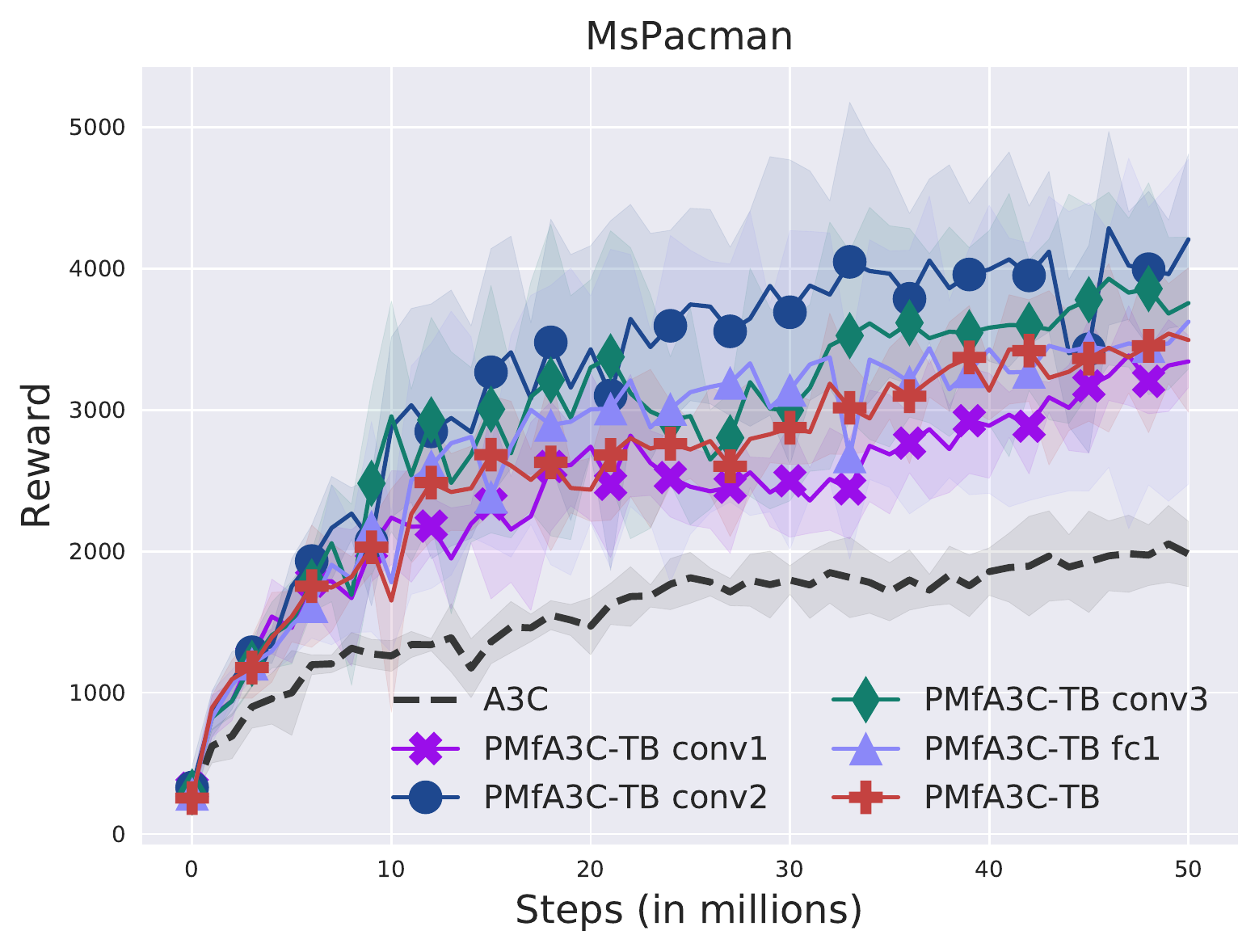}
    \end{subfigure}
    \begin{subfigure}[h]{\textwidth}
        \includegraphics[width=0.33\textwidth]{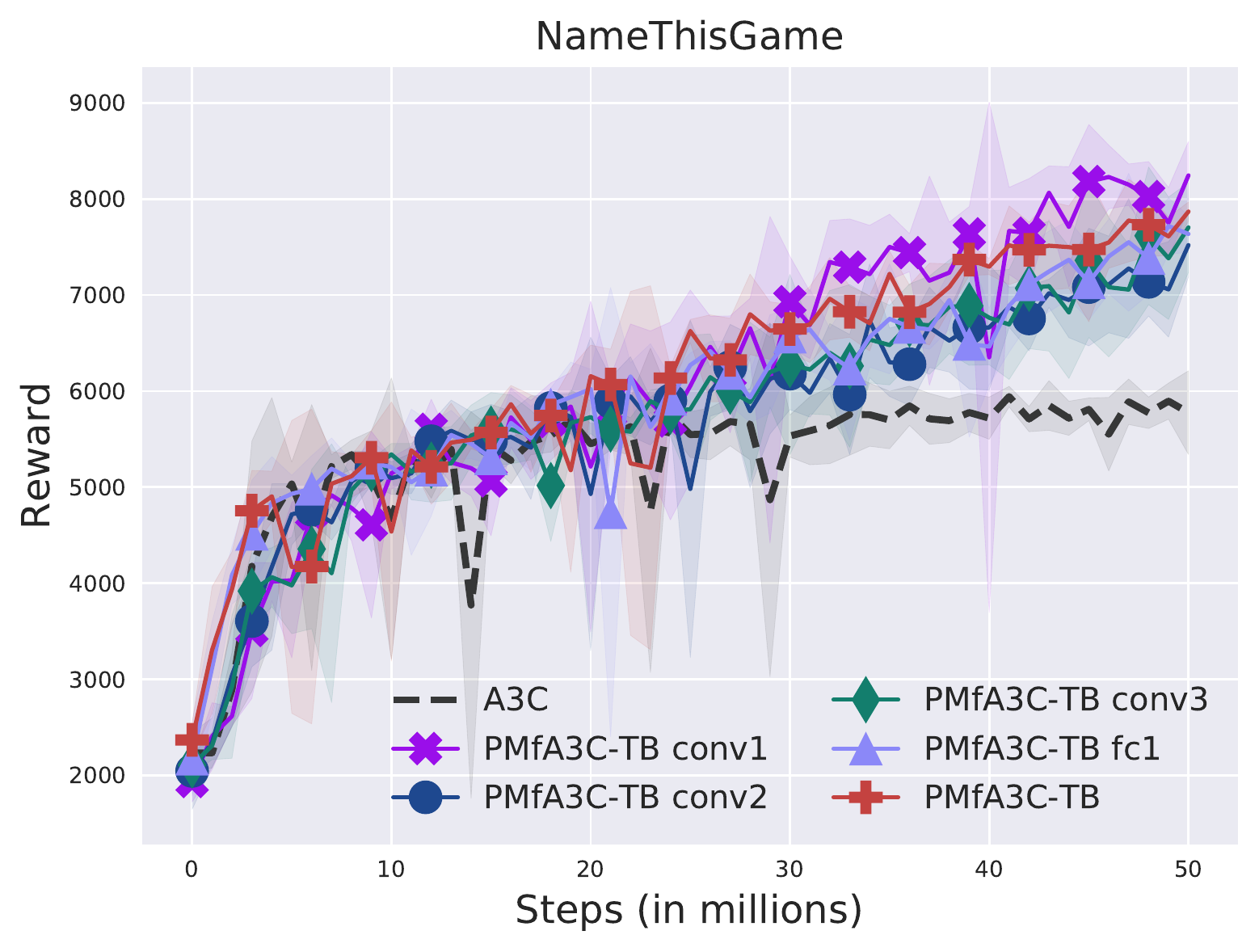}
        \includegraphics[width=0.33\textwidth]{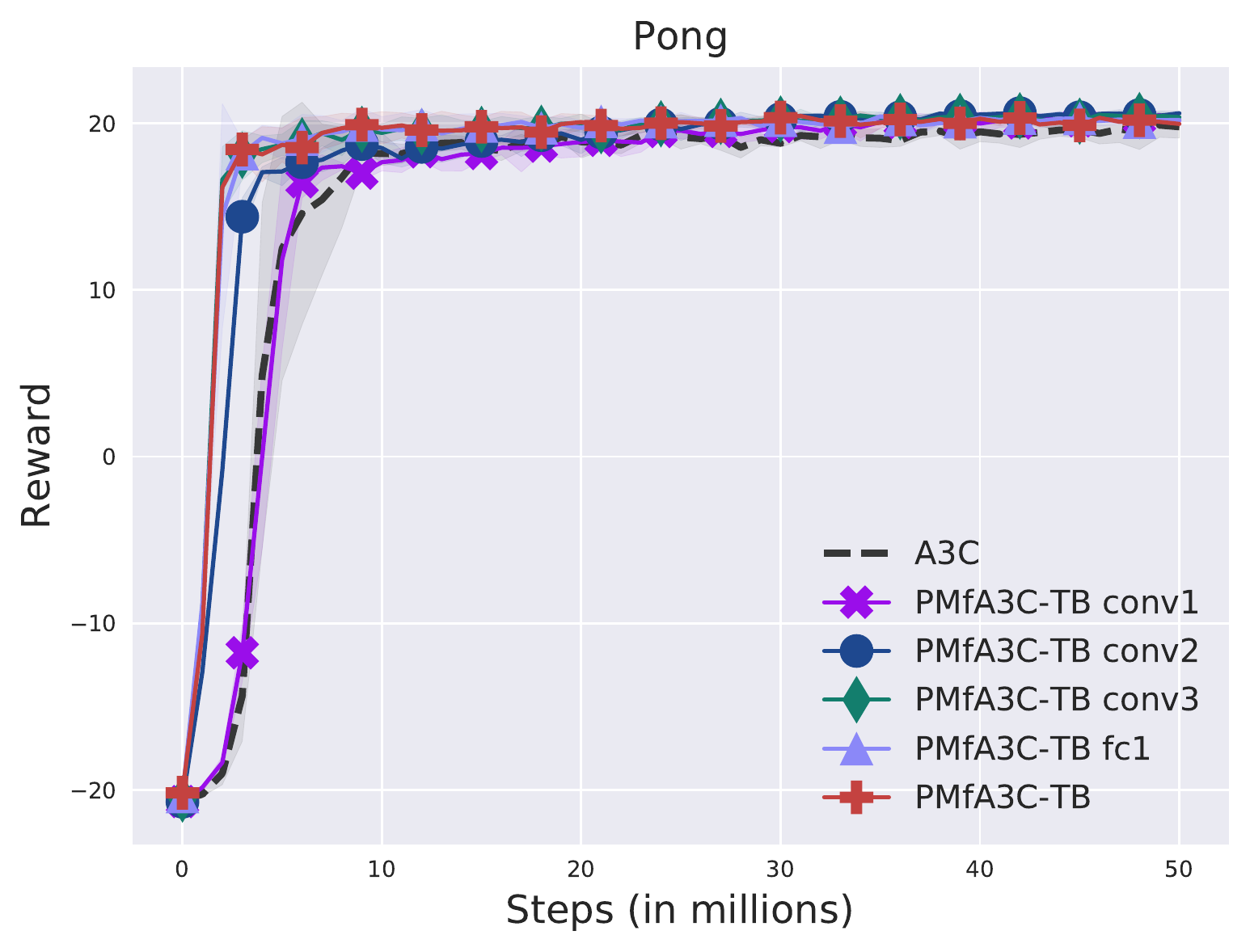}
        \includegraphics[width=0.33\textwidth]{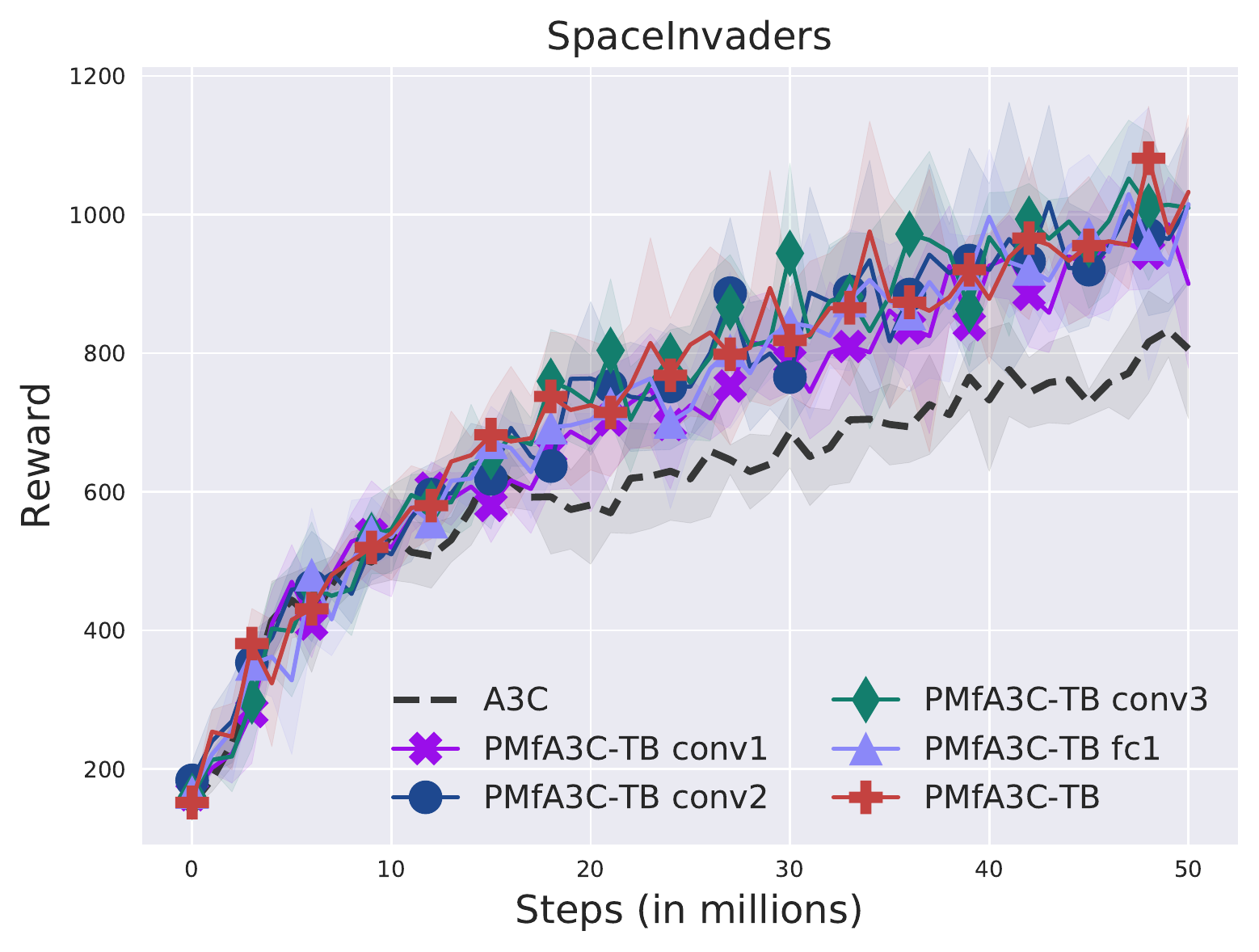}
    \end{subfigure}

    \caption{Performance of baseline and pre-training using A3C. The x-axis is the total number of training steps (among all $16$ actors), where each step consists of four game frames (we use frame skip of four). The y-axis is the average testing score over four trials where the shaded regions correspond to the standard deviation.}
    \label{fig:a3c_transfer_result}
\end{figure}

To understand how reusing just a subset of the whole pre-trained network affects the performance speedup and how it compares to our best approach PMfA3C-TB, we conduct the following ablation studies:
\begin{itemize}
    \item \emph{PMfA3C-TB}: reuse all layers (as was done in Section \ref{sec:pretrain-results}).
    \item \emph{PMfA3C-TB fc1}: reuse conv1, conv2, conv3, and fc1.
    \item \emph{PMfA3c-TB conv3}: reuse conv1, conv2, and conv3.
    \item \emph{PMfA3c-TB conv2}: reuse conv1 and conv2. 
    \item \emph{PMfA3c-TB conv1}: reuse conv1 only.
\end{itemize}
Figure~\ref{fig:a3c_transfer_result} shows the results for our ablation study. We found the most intuitive results in Pong where the more pre-trained layers are reused, the better the results are. In Breakout and SpaceInvaders, reusing different layers show no obvious distinctions in performance compared to reusing all layers. In Asterix, reusing all pre-trained layers still has the best performance; when reusing only a part of the pre-trained layers, the performance is similar to the baseline A3C. MsPacman and NameThisGame have the most varying results when different layers are reused. It is interesting to note that they show inverted performance for different reusing strategies---PMfA3C-TB conv1 has the worst result in MsPacman among all reusing strategies but shows the best result in NameThisGame. Despite this distinction, all pre-training methods in these two games outperform the baseline, regardless of the choice of reuse layers.  

In summary, our ablation study shows that reusing the entire pre-trained network consistently performs well---it either achieves the best results or is comparable to reusing other layers.
We point out that reusing layers that are closer to the output layer (PMfA3C-TB fc1 and conv3) also shows competitive results in four out of six games. This is consistent with the findings in \cite{yosinski2014transferable} that lower layers learn general features while higher layers learn task-specific features, and transferring general features are more beneficial when the training and testing data are different. In our case, however, since the classification model and the A3C network are trained on data that are collected from the same game, leveraging task-specific features can be more helpful than reusing only general features.

\subsection{What is really learned from pre-training?}
In order to understand the benefit of our pre-training method, we visualize the feature maps in the last convolutional layer to assess if an RL agent can learn meaningful high-level information about the game through pre-training. We adopt the Gradient-weighted Class Activation Mapping (Grad-CAM) as our visualization method since it is model-agnostic and can be used in any convolutional-based network without needing to change architectures \citep{selvaraju2017grad}. Using Grad-CAM, we analyze how different regions in feature maps are activated by corresponding actions under three settings: 1) a randomly initialized RL agent, 2) a pre-trained classification model, and 3) a final learned RL agent in PMfA3C-TB. By comparing feature map patterns among the three scenarios, we will be able to obtain an intuitive understanding of what is learned through pre-training and how an RL agent uses the pre-trained knowledge during its learning. 

Grad-CAM is a visualization tool designed to increase the interpretability of prediction results of a deep neural network~\citep{selvaraju2017grad}. Specifically, Grad-CAM computes the gradient of the target logits (output values before applying softmax) %\gabe{modified}\MET{Define logit} 
with respect to feature maps of a convolutional layer. The importance weight of each neuron can then be obtained by performing global averaging pooling \citep{globalavgpool} over the gradients. For a classification task, the target logits refer to the score of a target class, while in the context of RL, the target class can be considered as the action $a$ taken based on the current policy $\pi$. Given any game state $s$, the score $y^a$ of an action $a$ is defined as $y^a = \pi(a|s;\theta)$, where $\theta$ parameterizes the policy network in A3C. We then compute gradients for $y^a$ with respect to the feature maps of the last convolutional layer (denoted as $M^k$).\footnote{In \cite{selvaraju2017grad} the feature maps are denoted as $A^k$, we change this notation to avoid confusion with the notation of the action $a$ in the context of RL.} We compute the importance weight (denoted as $\alpha_{k}^{a}$) in A3C as
\[\alpha_{k}^{a} = \overbrace{\frac{1}{Z} \sum_{i} \sum_{j}}^\text{global average pooling} \underbrace{\frac{\partial y^{a}}{\partial M^{k}_{ij}}}_\text{gradients via backprop}\]
\cite{selvaraju2017grad} also combines forward activation maps and pass through a rectifier linear units (ReLU) to show the features that has a \emph{positive} effect on the action of interest. This gives us the action-discriminative saliency map as
\[L^{a}_{\text{Grad-CAM}} = ReLU\left(\sum_{k}\alpha^{a}_{k}M^{k}\right)\]
However, the output of the discriminative saliency map did not provide an interpretable visualization under A3C. 
According to \citep{selvaraju2017grad}, the weight $\alpha^{a}_{k}$ captures the \emph{importance} of the feature map $k$. Since $\alpha^{a}_{k}$ can have both positive and negative values, we choose to emphasize only on positive weights (i.e., captures the most important features).
%\gabe{modified}\MET{I don't see how the previous statement implies the following sentence.} 
Thus, we instead apply ReLU directly to $\alpha^{a}_{k}$, transforming $L^{a}_{\text{Grad-CAM}}$ as
\[
L^{a}_{\text{Grad-CAM}} = \sum_{k}ReLU\big(\alpha^{a}_{k}\big)M^{k}    
\]

\begin{figure}[pt!]
  \centering
    \includegraphics[width=\textwidth]{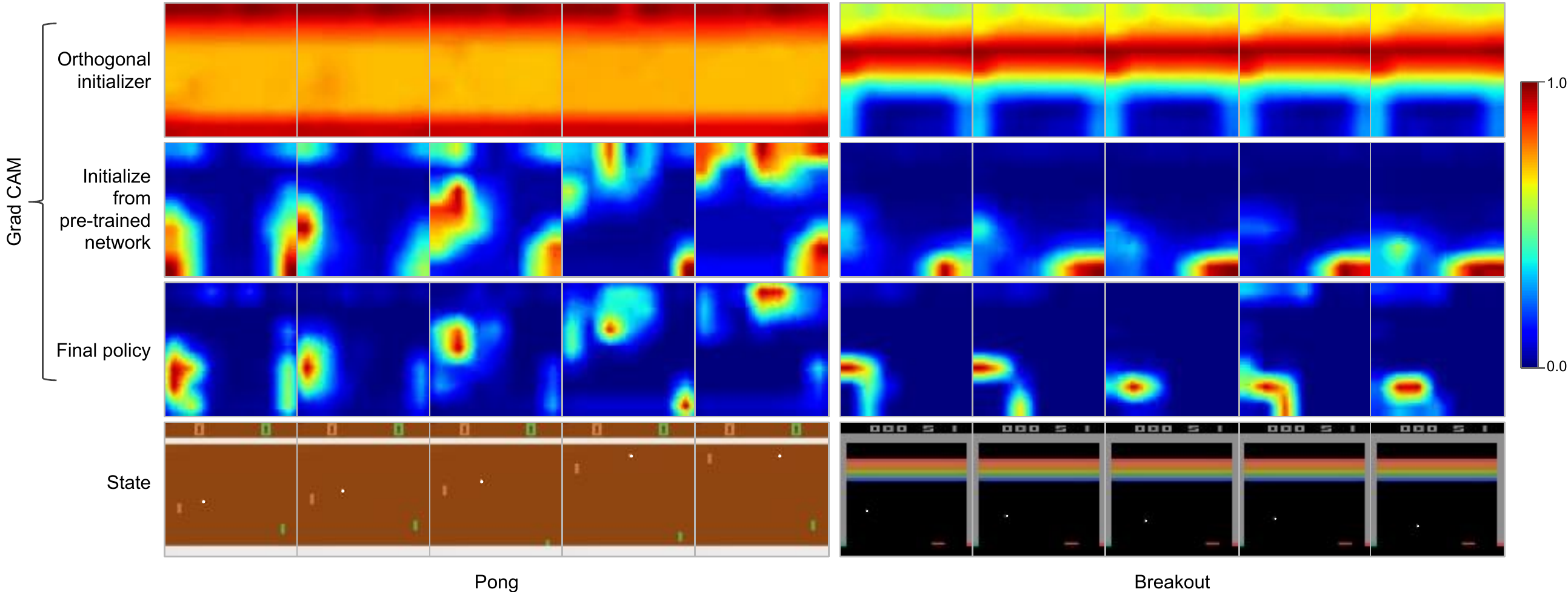}
    \caption{Gradient-weighted Class Activation Mapping (Grad-CAM) visualization for Pong and Breakout. Top row: random initialization; second row: pre-trained model; third row: final learned RL policy; bottom row: the original game image.}
    \label{fig:grad_cam}
\end{figure}

We present Grad-CAM results for two games, Pong and Breakout, as the running example in this section for analyzing what features are learned. Video clips for all six games' Grad-CAM results are available online at \url{https://sites.google.com/view/pretrain-deeprl}. Figure \ref{fig:grad_cam} shows example Grad-CAM results for a sequence of five frames in Pong and Breakout and each frame is used as the input image to the network. We run a forward pass to compute the target logits $y^a$ and output an action $a$ based on the current policy $\pi$. Note that actions are not executed in the environment but only used to compute $L^{a}_{\text{Grad-CAM}}$---we do not perform gradient updates to the network. To ensure we compare the three models (random initialized, pre-trained, and final learned RL) with the same set of game inputs, we perform one episode of evaluation (until the game ends or reaches 5,000 testing steps, whichever comes first) using the final learned RL agent and save all game images it encounters, then use these images as a sequential state input to calculate Grad-CAM for the random initialized agent and the pre-trained model. Grad-CAM is visualized as a heatmap of scale $[0, 1]$: red (value close to $1$) indicates regions that are important for a corresponding action while blue (value close to $0$) indicates unimportant regions. The top row shows the feature map for a randomly initialized agent; the second row is the feature map of a pre-trained classifier; the third row shows a final learned RL agent's feature map; the bottom row shows the original game image. 

At the beginning of training when the network weights are randomly initialized (top row of Figure \ref{fig:grad_cam}), both Pong and Breakout agents act randomly and overall consider the entire game state to be important. In particular, Pong agent sees the top part (where the score is located shows more red color) and somehow the bottom part to be more important than the middle part (orange color) of the game image; Breakout agent also considers the top part to be important---with more focus on the location of bricks (red-orange color) than the location of the score (yellow-green color)---but the bottom part of the image is shown to be not important at all (blue color). 

The important regions change notably when the network reuses weights from pre-trained models (second row of Figure \ref{fig:grad_cam}). Interestingly, in both games, the agent learns to pay attention to the paddle movements. In Pong, both the opponent's paddle (left) and the agent's paddle (right) are identified to be important; in Breakout, the paddle at the bottom becomes the most important (whereas under random initialization the bottom part of the image is considered to be the least). The result is intuitive for the pre-trained model as it finds the object that follows the cardinal directions of the actions taken to be the most important feature---when the demonstrator takes an action, the object that correlates the most to this action is the paddle. 

We see a further change of important regions in the final learned RL policy (third row of Figure \ref{fig:grad_cam}). Instead of paying attention to the paddle, both games learn to track the movement of the ball. This is particularly clear in Pong: the important features evolve following roughly the same trajectory as the movement of the ball, from the bottom-left to the top-right of the game image. 

While the Grad-CAM examples in Figure \ref{fig:grad_cam} show that the pre-trained model picks up different important regions than the final RL policy, we also observe that some pre-training features are carried on to the final RL policy. For example in Pong, although the RL policy has more focus on the ball, the regions around paddles are still activated with a lower importance---an inheritance from the pre-train model. In addition, in the Grad-CAM video clips where we show a complete run on all six games, it is more clear that important features in the pre-training are also identified in the final RL policy. We refer the readers to the video clips at \url{https://sites.google.com/view/pretrain-deeprl}.  

\section{Conclusion and Discussion}
\label{sec:conclusion}
The goal of this article is not to defeat the state-of-the-art results for Atari. Instead, we want to address the problem of slow learning. Other work focuses on increasing computational resources to speed up the training, while our work focuses more on improving learning speed, which is the ability to learn better policy with a smaller amount of game environment interactions. We attain the speedup in learning by supervised pre-training of the deep RL's network using non-expert human demonstrations.

We used the transformed Bellman operator~\citep{pohlen2018observe} in A3C, which addresses reward clipping that allows the RL agent to differentiate high and low rewarding states. The transformed Bellman operator helps our pre-training approach achieve improvements in all six games that we evaluated our approach. Our pre-training approach improves the reward in MsPacman with $67.36\%$. In addition, A3C in MsPacman reach the highest average reward of 2,052 at 49 million training steps, while PMfA3C-TB in MsPacman only takes 11 million training steps to surpass A3C's highest average reward. This is quite a significant speedup in the learning speed especially when we only pre-trained PMfA3C-TB's network on 14,504 game states from a non-expert human. 
%Where one game state and one training step is equivalent to a single interaction with the environment. 

We also have a much better understanding of what features are being considered during pre-training by using our modified Grad-CAM. One future direction for pre-training in deep RL is to drive the activation mapping towards the objects in the game state image. We believe it would be easier for non-expert humans to simply identify the important objects in the game relative to actually playing the game---this would be another way of using humans to improve learning.

As we investigate further ways to improve our approach, we know there is a limit to how much improvement pre-training can provide without addressing policy learning. In our approach, we have already trained a model with a policy that tries to imitate the human demonstrator, and thus we can extend this work by using the pre-trained model's policy to provide advice to the agent~\citep{wang2017improving}.

To summarize, learning both features and policy directly from raw images through deep neural networks is a major factor why learning is slow in deep RL. This article has demonstrated the following: 1) we have shown through Grad-CAM that using supervised pre-training with non-expert human demonstration data can be used for feature learning, and 2) that our method of initializing deep RL's network with a supervised pre-trained model can significantly speed up learning in deep RL.

\acks
The A3C implementation was a modification of \url{https://github.com/miyosuda/async_deep_reinforce}. The authors thank Sahil Sharma and Kory Matthewson for providing very useful insights on the actor-critic method. We also thank NVidia for donating a graphics card used in these experiments. This research used resources of Kamiak, Washington State University's high performance computing cluster, where we ran all our experiments.

% \MET{Abtahi: h3 = ?}
% \MET{Kempka: doom -> Doom, ai -> AI}
% \MET{Kurin: atari -> Atari}
% \MET{Papoudakis: doom -> Doom}
% \MET{Pohlen: atari -> Atari}
% \MET{Vinyals: Starcraft ii -> Starcraft II}
% \gabe{references fixed, Abathi h3 is the page number so it should be correct}

\bibliographystyle{agsm} 
\bibliography{KER}

\end{document}